\documentclass[10pt,twocolumn,letterpaper]{article}

\usepackage[pagenumbers]{wacv} 

\usepackage{graphicx}
\usepackage{caption}
\usepackage{multirow}
\usepackage[table]{xcolor}
\usepackage[ruled,vlined, linesnumbered]{algorithm2e} 
\usepackage{pifont}
\usepackage{tikz}
\newcommand*\circled[1]{\tikz[baseline=(char.base)]{
            \node[shape=circle,fill,inner sep=.2pt] (char) {\textcolor{white}{#1}};}}

\usepackage[most]{tcolorbox}
\usepackage{lipsum}
\tcbset{
  myquote/.style={
    colback=teal!5!white,
    colframe=teal!60!black,
    boxrule=0.5pt,
    arc=4pt,
    left=1pt, right=1pt,
    top=2pt, bottom=2pt,
    fonttitle=\bfseries,
    halign=center,
    enhanced
  }
}

\definecolor{DarkYellow}{rgb}{0.85, 0.65, 0.13}

\newcolumntype{M}{@{\hspace{.3em}}c@{\hspace{.3em}}}

\definecolor{wacvblue}{rgb}{0.21,0.49,0.74}
\usepackage[pagebackref,breaklinks,colorlinks,allcolors=wacvblue]{hyperref}

\def\wacvPaperID{857} 
\def\confName{WACV}
\def\confYear{2026}

\title{UI-Styler: Ultrasound Image Style Transfer with Class-Aware Prompts for Cross-Device Diagnosis Using a Frozen Black-Box Inference Network
\vspace{-.5em}
}

\author{
Nhat-Tuong Do-Tran \qquad Ngoc-Hoang-Lam Le \qquad Ching-Chun Huang \\
National Yang Ming Chiao Tung University \\
{\tt\footnotesize $\text{\{tuongdotn.ee12, lengochoanglam.ee12, chingchun\}}$@nycu.edu.tw}\\
{\small \url{https://dotrannhattuong.github.io/UIStyler}}
}

\begin{document}
\maketitle

\begin{abstract}
The appearance of ultrasound images varies across acquisition devices, causing domain shifts that degrade the performance of fixed black-box downstream inference models when reused. To mitigate this issue, it is practical to develop unpaired image translation (UIT) methods that effectively align the statistical distributions between source and target domains, particularly under the constraint of a reused inference-blackbox setting. However, existing UIT approaches often overlook class-specific semantic alignment during domain adaptation, resulting in misaligned content-class mappings that can impair diagnostic accuracy.
To address this limitation, we propose UI-Styler, a novel ultrasound-specific, class-aware image style transfer framework. UI-Styler leverages a pattern-matching mechanism to transfer texture patterns embedded in the target images onto source images while preserving the source structural content. In addition, we introduce a class-aware prompting strategy guided by pseudo labels of the target domain, which enforces accurate semantic alignment with diagnostic categories.
Extensive experiments on ultrasound cross-device tasks demonstrate that UI-Styler consistently outperforms existing UIT methods, achieving state-of-the-art performance in distribution distance and downstream tasks, such as classification and segmentation.
\vspace{-1.5em}
\end{abstract}

\section{Introduction} \label{sec:intro}
In ultrasound medical applications, downstream models (DMs) are typically trained on a specific domain (i.e., the target device) and often experience performance degradation when applied to a different domain — a phenomenon known as domain shift~\cite{domainshift_2, ECB, HiGDA, DARK}. Fully fine-tuning DMs for each new domain is generally impractical, as it is both time-consuming and resource-intensive. To mitigate this, prompt-tuning (PT) protocols~\cite{vpt, vpt2, promptcam, E2vpt} have been proposed, which adapt DMs or large-scale foundation models (LFMs) to new domains by modifying the input space or internal representations using a small number of learnable prompt parameters. More recently, \textbf{gradient-free prompt methods}~\cite{blackvip, CraFT, BAPs} have been introduced to enable adaptation without accessing backbone parameters, making them suitable for scenarios where downstream models are treated as black-box models (BMs) and accessed only through APIs (i.e., as in our setting). However, despite their success in computer vision tasks, these methods still require annotated data, limiting their applicability in fully unsupervised settings.

\begin{figure}[!t]
\centering
\includegraphics[width=0.9\linewidth]{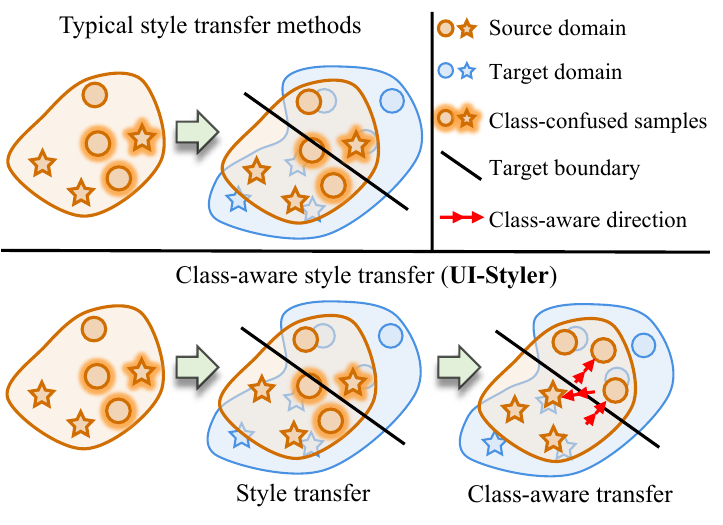}
\vspace{-.8em}
\caption{
Comparison between the typical \textbf{unpaired} image style transfer methods (top) and our proposed class-aware style transfer approach (bottom) for cross-device ultrasound diagnosis. Conventional methods align source and target distributions at the domain level but often neglect class-level alignment, leading to \emph{misaligned mappings}, especially for unlabeled (class-confused) samples. In contrast, \textbf{UI-Styler} enforces class-aware alignment via class-specific prompting, guiding class-confused samples toward their \emph{correct semantic classes}. The target class boundary reflects the behavior of the frozen black-box inference network.
} \label{fig:1}
\vspace{-1.2em}
\end{figure}

To address this problem, prompt-based domain adaptation (PDA) methods~\cite{vdpg, l2c} have leveraged prompt learning strategies to guide BMs' features toward the target domain. However, both PT and PDA approaches encounter two key limitations when applied to medical ultrasound data:
\circled{1} They rely heavily on the generalization capability of BMs — a requirement that is \textbf{rarely} met in small-scale ultrasound datasets. As shown in \cref{tab:dataset_sizes}, even the relatively large medical dataset BUSBRA~\cite{BUS-BRA} is more than $640\times$ smaller than the small web-scale dataset, ImageNet-1K~\cite{ImageNet}.
\circled{2} They assume logits or intermediate features are accessible from BMs, which is not feasible in commercial deployment scenarios where only the final predictions are available.


We refer to this scenario as the \textbf{inference-blackbox setting}, where the \textit{black-box downstream model}, pre-trained on the target domain, is frozen—without access to its parameters, gradients, intermediate features, or logits—and only provides final predictions. In this setting, only source and target data (e.g., images acquired from two different devices) are available, without any labels or paired information.
Note that in ultrasound imaging, appearance variations across acquisition devices pose challenges for a black-box model adapting to unfamiliar scanners.
Motivated by these observations, we pose the following open question:
\vspace{-.5em}
\begin{tcolorbox}[myquote]
How can we transfer the appearance of ultrasound images to align with the diagnostic behavior of the black-box downstream model?
\end{tcolorbox}
\vspace{-.5em}

For this, unpaired image translation (UIT) methods \cite{StyTr2, s2wat, USGAN} have emerged as promising alternatives for bridging cross-device appearance gaps by mapping a source image $I^{s-style}_{\textcolor{red}{s-content}}$ to its target-style counterpart $I^{\textcolor{teal}{t-style}}_{\textcolor{red}{s-content}}$ using the target images $I^{\textcolor{teal}{t-style}}_{t-content}$ as style reference. Although existing UIT methods effectively transfer image-level distributions between domains, they often overlook class-level information. As illustrated at the top of~\cref{fig:1}, naive style transfer can result in semantic misalignment, producing class-confused samples. In other words, without explicit class guidance, source representations may lose their discriminative characteristics during translation.

\textbf{Motivation.} To answer the above question, we propose \textbf{UI-Styler}, a class-aware style transfer framework specifically designed for unpaired and unsupervised settings—where neither ground-truth labels nor paired information is available for source and target samples—under an inference-blackbox reusage constraint. As illustrated in the bottom of~\cref{fig:1}, UI-Styler is engineered to achieve two primary objectives: (1) to mitigate domain-level appearance discrepancies by transferring source images to align with the target domain’s style, and (2) to preserve class-discriminative semantics by aligning source representations with class-specific structures implicitly captured by the frozen black-box inference network in the target domain. To achieve these objectives, UI-Styler adopts a dual-level stylization mechanism. At the domain level, it employs a cross-attention strategy to adapt source features to target style patterns while retaining the source’s structural content. At the category level, we introduce a novel class-aware prompting strategy that incorporates additional class-specific information into the stylized features (i.e., extracted by the style transfer step), with the goal of generating stylized images that accurately express their class characteristics. These prompts, learned from pseudo target labels, guide the stylized source features toward their correct semantic regions in the target domain. In essence, the learned prompts capture inter-class distinctions and approximate the normal directions of the decision boundaries present in the target domain, effectively steering the class-aware stylization process.

\textbf{Contributions.} Our main contributions are as follows:
\begin{enumerate}
\item We propose UI-Styler, which performs style transfer from the source to the target domain under an unpaired and unsupervised cross-domain setting, facilitating the reuse of a frozen, black-box downstream model.
\item We propose a dual-level stylization mechanism that adapts source images to the target domain via a pattern-matching approach for domain-level appearance and a class-aware prompting strategy, informed by the black-box downstream model, for class-level alignment.
\item Extensive experiments on $12$ cross-device tasks show that UI-Styler achieves state-of-the-art stylization performance in distribution distance and downstream task evaluation, including classification and segmentation.
\end{enumerate}

\begin{table}[!t]
\centering
\resizebox{.75\columnwidth}{!}{%
\begin{tabular}{l|l|c}
\toprule
Dataset Type & Dataset & \#Samples \\
\midrule
\multirow{4}{*}{Ultrasound} 
    & BUSI~\cite{BUSI}   & 647 \\
    & UCLM~\cite{UCLM}   & 264 \\
    & UDIAT~\cite{UDIAT} & 163 \\
    & BUSBRA~\cite{BUS-BRA} & 1{,}875 \\
\midrule
\multirow{3}{*}{Web-scale} 
    & ImageNet-1K~\cite{ImageNet}     & 1.2M \\
    & ImageNet-21K~\cite{ImageNet-21K} & 12.7M \\
    & CLIP's dataset~\cite{clip}       & 400M \\
\bottomrule
\end{tabular}
}
\vspace{-.8em}
\caption{Comparison of the number of samples across ultrasound datasets and web-scale datasets. ``M" denotes millions of samples.} \label{tab:dataset_sizes}
\vspace{-1.2em}
\end{table}

\begin{figure*}[!t]
\centering
\includegraphics[width=0.75\linewidth]{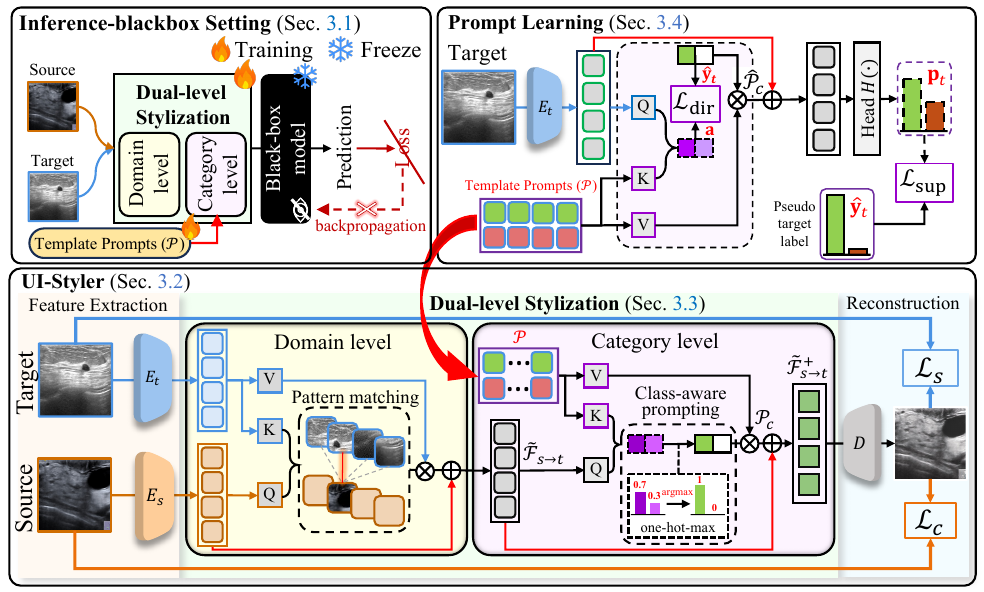}
\vspace{-.7em}
\caption{
\textbf{Top-left}: Overview of the proposed UI-Styler framework for ultrasound image translation under an inference-blackbox setting. Given unlabeled source and target images, UI-Styler performs dual-level stylization along with template prompt set $\mathcal{P}$. The black-box downstream model is frozen and is only for final predictions.
\textbf{Bottom}: Details of the dual-level stylization module (\cref{method:dual}). At the domain level, pattern matching is performed via cross-attention to inject target style into source content. At the category level, given the learned prompt set $\mathcal{P}$, a class-specific prompt $\mathcal{P}_c$ is determined and used to refine the stylized features $\widetilde{\mathcal{F}}_{s \rightarrow t}$. The final stylized image is reconstructed by a decoder $D$ and optimized using content and style losses ($\mathcal{L}_c$, $\mathcal{L}_s$). \textbf{Top-right}: The prompt set $\mathcal{P}$ is optimized using $\mathcal{L}_{\text{dir}}$ and $\mathcal{L}_{\text{sup}}$ (\cref{method:training}) to capture the distinctive characteristics of each semantic class as defined by the black-box model. Note that the encoder $E_t$ and the cross-attention network (highlighted in \textcolor{magenta}{pink}) share the same weights as those used in the UI-Styler model (bottom part).}
\vspace{-1.em}
\label{fig:2}
\end{figure*}

\section{Related Works}
\subsection{Unpaired Image Translation}
Unpaired image translation (UIT) aims to map images from a source domain to the visual style of a target domain without requiring paired supervision. Early UIT methods~\cite{SANet, adain, adaattn} employed convolutional encoder-decoder architectures \cite{CNN} to align domain distributions, but they were limited in capturing long-range dependencies, often producing stylized images lacking fine details. Moreover, as maintaining tissue structure is a critical property in ultrasound imaging for accurate diagnosis, transformer-based approaches \cite{StyTr2, s2wat, tanrscolour} have emerged, leveraging their ability to model global context and preserve structural information. For instance, StyTr$^2$ \cite{StyTr2} employs a dual-encoder Vision Transformer (ViT) \cite{ViT} with content-aware positional encoding to capture precise content representations and preserve fine-grained details during stylization.
Similarly, US-GAN \cite{USGAN} adapts UIT specifically for ultrasound image translation by decomposing latent features into content and texture components to enable fine-grained texture transfer while maintaining structural consistency. Even so, most prior works primarily focus on mitigating domain-level shifts while neglecting class-level semantics, which can lead to class ambiguity in the translated images. To address this issue, our proposed UI-Styler refines stylized features to align not only with the target domain style but also with class-discriminative semantics through a class-aware prompting mechanism.

\subsection{Prompt Tuning}
Prompt tuning \cite{vpt, vpt2} has emerged as a parameter-efficient alternative to full model fine-tuning for adapting large-scale foundation models to new tasks. By injecting learnable prompts at the input or intermediate layers, it enables control over model behavior with minimal trainable parameters. Building on this paradigm, gradient-free prompt tuning methods \cite{blackvip, CraFT, BAPs} extend to black-box settings, where access to model parameters is restricted, making them suitable for API-based downstream models (DMs). However, these approaches still assume the availability of much labeled data, which is often costly and impractical. 

To address both annotation scarcity and black-box constraints, recent studies \cite{vdpg, l2c} have explored prompt-based domain adaptation, which guides DMs by consolidating their input or output space through domain-specific prompts. Yet, these methods typically rely on large-scale labeled datasets to train prompts prior to deployment and assume that DMs expose intermediate features or logits (e.g., as in CLIP \cite{clip}). This assumption often does not hold in commercial DMs or privacy-sensitive scenarios, where only \textbf{DM's final predictions} are accessible—a situation known as the \textbf{inference-blackbox setting}. In contrast, our work targets this underexplored setting, where \textbf{no} labels, gradients, or DM's features are available—particularly relevant to medical applications, where large-scale labeled datasets are infeasible and reusing the DMs is essential.

\section{Methodology} \label{sec:method}
In this section, we present the proposed UI-Styler framework for unpaired and unsupervised style transfer under an inference-blackbox setting, as illustrated in~\cref{fig:2}. We begin by formally defining the problem in~\cref{method:problem} and then provide an overview of the overall architecture in~\cref{method:overview}. Subsequently, we detail the core dual-level stylization module in~\cref{method:dual}, followed by a description of the training strategy in~\cref{method:training}.

\subsection{Problem Setting} \label{method:problem}
We consider the problem of unpaired and unsupervised style transfer under an inference-blackbox setting, aiming to translate source ultrasound images to match the target domain's style while preserving diagnostic semantics. Let $\mathcal{D}_s = \{x_s^i\}_{i=1}^{N_s}$ denote the \emph{source domain}, containing $N_s$ unlabeled ultrasound images $x_s^i \in \mathbb{R}^{H \times W \times 3}$ from a specific acquisition device. Conversely, the \emph{target domain} $\mathcal{D}_t = \{(x_t^j, \hat{y}_t^j)\}_{j=1}^{N_t}$ consists of $N_t$ ultrasound images $x_t^j \in \mathbb{R}^{H \times W \times 3}$ accompanied by pseudo labels $\hat{y}_t^j \in \mathcal{Y}$ generated by a black-box downstream model (BDM). Since the ground-truth (GT) labels for the source and target images are not available, we consider our setting unsupervised. Furthermore, we assume there is no paired correspondence between the source and target samples (i.e., $\mathcal{D}_s \cap \mathcal{D}_t = \emptyset$). \textbf{Importantly}, our method does not require access to BDM's parameters~\cite{vpt}, extracted features~\cite{vdpg}, or intermediate logits, making it well-suited for inference-blackbox scenarios.


\subsection{Architecture Overview} \label{method:overview}

The proposed end-to-end UI-Styler framework, as illustrated at the bottom of ~\cref{fig:2}, consists of three main modules: feature extraction, dual-level stylization with template prompts, and image reconstruction.

\textbf{Firstly}, given source and target images $x_s, x_t$, we extract visual features using two distinct Vision Transformer (ViT) encoders~\cite{ViT}: a source encoder $E_s$ and a target encoder $E_t$. As a result, the source and target features are defined as:
\[
\mathcal{F}_s = E_s(x_s) \in \mathbb{R}^{L \times d}, \quad
\mathcal{F}_t = E_t(x_t) \in \mathbb{R}^{L \times d},
\]
where $L = h\times w$ with $h = H / P$ and $w = W / P$ 
are the spatial dimensions corresponding to a patch size of $P \times P$, and $d$ denotes the embedding dimension of a patch token.

\textbf{Next}, our proposed dual-level stylization module narrows both \ding{172} domain-level and \ding{173} category-level discrepancies between the source and target datasets. \ding{172}~Pattern-matching mechanism (PM) transforms the source domain toward the target domain by integrating relevant style features $\mathcal{F}_t$ into the content representations $\mathcal{F}_s$, resulting in stylized features $\widetilde{\mathcal{F}}_{s \rightarrow t} \in \mathbb{R}^{L \times d}$.
\ding{173}~To address class ambiguity, class-aware prompting (CP) drives $\widetilde{\mathcal{F}}_{s \rightarrow t}$ toward class-specific distributions by leveraging the correlation between the $c$-th class prompt $\mathcal{P}_c \in \mathbb{R}^{L \times d}$ and the stylized features, resulting in class-aligned representations $\widetilde{\mathcal{F}}^+_{s \rightarrow t} \in \mathbb{R}^{L \times d}$. Here, these prompts serve as prototypical characteristics (e.g., benign tumors typically exhibit well-defined boundaries, whereas malignant ones tend to appear more blurred) and are learned using the pseudo labels of their target samples, as illustrated in the top-right of~\cref{fig:2}.

\textbf{Finally}, we reconstruct the stylized image $\widetilde{x}_s = D(\widetilde{F}^+_{s \rightarrow t}) \in \mathbb{R}^{H \times W \times 3}$ using a lightweight decoder $D$ composed of upsampling and convolutional layers \cite{StyTr2, SANet}.

\subsection{Dual-level Stylization} \label{method:dual}
Our dual-level stylization module follows a \textit{local-to-global} alignment principle, where \textbf{local} refers to token-level style adaptation through a pattern-matching mechanism, and \textbf{global} refers to feature-level semantic alignment via class-aware prompting. In this way, source representations are gradually transformed to align with both the visual appearance and semantic structure of the target domain, thereby enhancing downstream performance and improving physicians’ diagnostic capability on the source domain.

\noindent \textbf{Pattern-matching Mechanism.}
To align source content with target style, we adopt a cross-attention mechanism~\cite{attention, cross-att} that enables each source token to selectively incorporate the most relevant style patterns from the target domain. Specifically, the source-content features $\mathcal{F}_s$ are projected into queries, while the target-style features $\mathcal{F}_t$ are projected into keys and values:
\begin{equation}
\widetilde{\mathcal{F}}_{s \rightarrow t}^{(h)} = \mathrm{softmax}\left(\frac{Q^{(h)} {K^{(h)}}^\top}{\sqrt{d_h}}\right)V^{(h)},
\end{equation}
where $Q^{(h)} = \mathcal{F}_s W_q^{(h)}$, $K^{(h)} = \mathcal{F}_t W_k^{(h)}$, $V^{(h)} = \mathcal{F}_t W_v^{(h)}$, and $W_q^{(h)}, W_k^{(h)}, W_v^{(h)} \in \mathbb{R}^{d \times d_h}$ are learnable projection matrices for the $h$-th head. Here, $d_h$ denotes the dimensionality of each attention head and $\widetilde{\mathcal{F}}_{ s \rightarrow t}^{(h)}$ is the residual for stylization. The residual outputs from all heads are concatenated as $\left[ \widetilde{\mathcal{F}}_{s \rightarrow t}^{(1)}, \ldots, \widetilde{\mathcal{F}}_{s \rightarrow t}^{(H)} \right]$. Then, the stylized features are obtained by adding the output back to the original source features, followed by Layer Normalization \cite{layernorm} LN($\cdot$):
\begin{equation}
\widetilde{\mathcal{F}}_{s \rightarrow t} = \text{LN}([\widetilde{\mathcal{F}}_{s \rightarrow t}^{(1)}, \ldots, \widetilde{\mathcal{F}}_{s \rightarrow t}^{(H)}] + \mathcal{F}_s) \in \mathbb{R}^{L \times d}.
\end{equation}

\noindent \textbf{Class-aware Prompting.}
To resolve class ambiguity in the target stylized features 
$\widetilde{\mathcal{F}}_{s \rightarrow t}$, we introduce a set of learnable template prompts $\mathcal{P} \in \mathbb{R}^{C \times L \times d}$, where $C$ denotes the number of semantic classes (e.g., benign and malignant). These learned prompts (detailed in~\cref{method:training}) act as class-specific templates that capture the distinctive patterns of each class within the target domain. To select the most appropriate class-specific prompt for a given stylized feature $\widetilde{\mathcal{F}}_{s \rightarrow t}$ from the learned prompt template set $\mathcal{P}$, we compute a correlation vector between $\widetilde{\mathcal{F}}_{s \rightarrow t}$ and $\mathcal{P}$. To enforce a one-to-one assignment, we apply a one-hot encoding to the correlation vector by selecting the maximum entry, thereby performing a hard selection from the $C$ prompts. The selected class-specific prompt $\mathcal{P}_c ~~\in \mathbb{R}^{L \times d}$ is determined by:
\begin{equation}
\mathcal{P}_c = \mathsf{\text{one-hot-max}}\left(\mathcal{E}_f(\widetilde{\mathcal{F}}_{s \rightarrow t}) \mathcal{E}_p(\mathcal{P})^\top\right) \mathcal{P},
\end{equation}
where $\mathcal{E}_f(\cdot)$ and $\mathcal{E}_p(\cdot)$ denote the feature and prompt embedders, respectively, both implemented using lightweight convolutional layers. Finally, by adding the selected class-specific prompt to the stylized features, we obtain the final class-aligned representation as follows and push each sample toward its class's prototype. 
\begin{equation}
\widetilde{\mathcal{F}}^+_{s \rightarrow t} = \widetilde{\mathcal{F}}_{s \rightarrow t} + \mathcal{P}_c \in \mathbb{R}^{L \times d}.
\end{equation}

\subsection{Training Strategy} \label{method:training}
\noindent \textbf{Prompt Learning and Losses.} 
Given the target features $\mathcal{F}_t$, the prompt set $\mathcal{P}$ is optimized by jointly minimizing a direction loss ($\mathcal{L}_{\text{dir}}$) and a supervised loss ($\mathcal{L}_{\text{sup}}$), both guided by pseudo target labels $\hat{y}_t$. 
We assume the black-box functions as an image classifier; the pseudo target labels correspond to the predicted class by the black-box downstream model. The class-specific prompt is then defined as $\hat{\mathcal{P}}_c = \mathbf{\hat{y}}_t \mathcal{P}$, where $\mathbf{\hat{y}}_t \in \{0,1\}^{C}$ is the one-hot vector of $\hat{y}_t$. Learned $\hat{\mathcal{P}}_c$ is expected to approximate the normal direction of the decision boundary (hyperplane) for class $c$.
Given a target sample of class $c$, its feature should exhibit a positive correlation with $\hat{\mathcal{P}}_c$, and adding $\hat{\mathcal{P}}_c$ to the feature should improve its classification confidence.
Note that in our experiments, the black-box downstream model may also output a segmentation mask, which is used to assess the impact of image stylization on segmentation performance; the mask is not utilized during the prompt learning process.


To realize this idea, we define the direction loss based on a one-hot classification objective, which encourages the target features (including $\mathcal{F}_t$ and the target-stylized feature $\widetilde{\mathcal{F}}_{s \rightarrow t}$) to align closely with the corresponding class-specific prompt. Let $\mathbf{a} = \mathrm{sigmoid}(\mathcal{E}_f(\mathcal{F}_t) \mathcal{E}_p(\mathcal{P})^\top) \in \mathbb{R}^{C}$ denote the class correlation vector for a target feature $\mathcal{F}_t$ to the prompt set $\mathcal{P}$. The direction loss is computed as:
\vspace{-.5em}
\begin{equation}
\mathcal{L}_{\text{dir}} = - \frac{1}{C} \sum_{c=1}^{C} \left[ \hat{y}_c \log a_c + (1 - \hat{y}_c) \log (1 - a_c) \right],
\vspace{-.2em}
\end{equation}
where $\hat{y}_c = 1$ if $c=\hat{y}_t$; otherwise, $\hat{y}_{c} = 0$ for $c \neq \hat{y}_t$, and $a_c$ is the $c$-th element of the correlation vector $\mathbf{a}$. Moreover, supervised cross-entropy loss is defined as:
\vspace{-.25em}
\begin{equation}
\mathcal{L}_{\text{sup}} = - \mathbf{\hat{y}}_t\cdot\log(\mathbf{p}_t),
\vspace{-.2em}
\end{equation}
where we add the selected class prompt $\hat{\mathcal{P}}_c$ to the target feature $\mathcal{F}_t$ along with a classifier head $H(\cdot)$ to produce class probabilities, $\mathbf{p}_t = \mathrm{softmax}(H(\mathcal{F}_t + \hat{\mathcal{P}}_c)) \in \mathbb{R}^{C}$.

\noindent \textbf{Final Objective Function.}
The objective for training the proposed \textbf{UI-Styler} and the class prompts combines the aforementioned prompt losses with the stylization losses. Following prior style transfer works~\cite{StyTr2, SANet, s2wat}, we employ a content loss $\mathcal{L}_c$ to encourage the stylized output to preserve structural information from the source, and a style loss $\mathcal{L}_s$ to align the output appearance with the target domain. \textbf{The total loss below jointly optimizes the parameters of the encoders ($E_s$, $E_t$), the dual-level stylization module, the decoder ($D$), the prompt set ($\mathcal{P}$), and the prompt classifier head ($H(\cdot)$)}:
\vspace{-.25em}
\begin{equation}
\mathcal{L}_{\text{total}} = \lambda_{\text{dir}}\mathcal{L}_{\text{dir}} + \lambda_{\text{sup}}\mathcal{L}_{\text{sup}} + \lambda_c\mathcal{L}_c + \lambda_s\mathcal{L}_s,
\vspace{-.2em}
\end{equation}
where $\lambda_{\text{dir}}$, $\lambda_{\text{sup}}$, $\lambda_c$, and $\lambda_s$ denote the loss weights. We set all weights to $1$ in experiments, supported by a sensitivity analysis on loss balancing. Notably, the formulations of $\mathcal{L}_c$ and $\mathcal{L}_s$ (Sec.~\textcolor{wacvblue}{D}), as well as the sensitivity analysis (Secs. \textcolor{wacvblue}{F.1} and \textcolor{wacvblue}{F.2}), are detailed in the supplementary material.

\begin{table*}[!t]
\centering
\resizebox{\textwidth}{!}{%
\begin{tabular}{@{\hspace{.1em}}c@{\hspace{.1em}}||@{\hspace{.1em}}c@{\hspace{.1em}}|@{\hspace{.1em}}M M M M M@{\hspace{.1em}}|@{\hspace{.1em}}c@{\hspace{.1em}}|@{\hspace{.1em}}M M M M M@{\hspace{.1em}}|@{\hspace{.1em}}c@{\hspace{.1em}}|@{\hspace{.1em}}M M M M M@{\hspace{.1em}}}
\toprule
\textbf{Method} & 
\textbf{Tasks} & KID\textcolor{red}{$\downarrow$} & Acc\textcolor{red}{$\uparrow$} & AUC\textcolor{red}{$\uparrow$} & Dice\textcolor{red}{$\uparrow$} & IoU\textcolor{red}{$\uparrow$} &
\textbf{Tasks} & KID\textcolor{red}{$\downarrow$} & Acc\textcolor{red}{$\uparrow$} & AUC\textcolor{red}{$\uparrow$} & Dice\textcolor{red}{$\uparrow$} & IoU\textcolor{red}{$\uparrow$} &
\textbf{Tasks} & KID\textcolor{red}{$\downarrow$} & Acc\textcolor{red}{$\uparrow$} & AUC\textcolor{red}{$\uparrow$} & Dice\textcolor{red}{$\uparrow$} & IoU\textcolor{red}{$\uparrow$} \\
\midrule

w/o ST & \multirow{5}{*}{\shortstack{BUSBRA\\$\downarrow$\\BUSI}} & 17.74 & 71.40 & 73.35 & \underline{83.99} & \underline{74.05} & \multirow{5}{*}{\shortstack{BUSBRA\\$\downarrow$\\UCLM}} & 28.48 & \underline{64.12} & \underline{67.80} & 81.66 & 71.01 & \multirow{5}{*}{\shortstack{BUSBRA\\$\downarrow$\\UDIAT}} & 13.81 & 55.95 & 64.29 & 84.76 & 75.71 \\

TransColor \cite{tanrscolour} & & \underline{11.32} & 73.18 & 74.63 & 80.85 & 70.36 & & \textbf{16.85} & 56.48 & 61.65 & 78.67 & 67.38 & & 12.53 & 59.50 & 63.40 & 84.67 & 75.55  \\

S2WAT \cite{s2wat} & & 12.47 & \underline{73.89} & \underline{75.23} & 82.84 & 72.69 & & 16.93 & 62.88 & 63.05 & \underline{81.73} & \underline{71.25} & & \underline{10.08} & \underline{63.94} & \underline{65.93} & 85.67 & 76.74 \\

Mamba-ST \cite{mamba-st} & & 15.36 & 72.47 & 71.85 & 82.48 & 72.33 & & 19.25 & 55.42 & 63.66 & 81.29 & 70.75 & & 12.99 & 60.57 & 64.14 & \textbf{86.10} & \underline{77.21} \\

UI-Styler & & \textbf{11.20} & \textbf{75.84} & \textbf{76.33} & \textbf{84.52} & \textbf{74.74} & & \underline{16.91} & \textbf{75.13} & \textbf{76.78} & \textbf{82.06} & \textbf{71.73} & & \textbf{9.14} & \textbf{72.47} & \textbf{71.52} & \underline{86.04} & \textbf{77.52} \\

\midrule

w/o ST & \multirow{5}{*}{\shortstack{BUSI\\$\downarrow$\\BUSBRA}} & 19.73 & 82.56 & \underline{87.30} & 82.41 & 73.37 & \multirow{5}{*}{\shortstack{BUSI\\$\downarrow$\\UCLM}} & 18.39 & \underline{65.64} & \underline{68.77} & 77.65 & 67.97 & \multirow{5}{*}{\shortstack{BUSI\\$\downarrow$\\UDIAT}} & 7.23 & \underline{73.33} & 73.16 & 79.53 & 70.61 \\

TransColor \cite{tanrscolour} & & 12.38 & 82.56 & 85.83 & 81.64 & 72.32 & & 17.25 & 64.10 & 65.02 & 77.71 & 67.90 & & 7.02 & 69.23 & 71.05 & \underline{80.41} & \underline{71.44} \\

S2WAT \cite{s2wat} &  & \underline{11.67} & 80.51 & 84.88 & \underline{82.85} & \underline{73.70} & & 15.61 & 62.56 & 57.38 & 77.35 & 67.45 & & \textbf{3.37} & 71.79 & \underline{73.38} & 80.06 & 71.02 \\

Mamba-ST \cite{mamba-st} & & 14.12 & \underline{84.62} & 86.58 & 81.53 & 72.30 & & \underline{15.11} & 65.13 & 63.93 & \underline{77.89} & \underline{68.15} & & 4.27 & 71.28 & 71.76 & 80.30 & 71.39 \\

UI-Styler & & \textbf{11.25} & \textbf{85.13} & \textbf{88.14} & \textbf{83.15} & \textbf{74.05} & & \textbf{11.05} & \textbf{74.36} & \textbf{77.15} & \textbf{78.83} & \textbf{68.61} & & \underline{3.61} & \textbf{74.36} & \textbf{78.89} & \textbf{80.49} & \textbf{71.61} \\

\midrule

w/o ST & \multirow{5}{*}{\shortstack{UCLM\\$\downarrow$\\BUSBRA}} & 26.74 & \underline{87.50} & \underline{92.29} & 81.68 & 71.73 & \multirow{5}{*}{\shortstack{UCLM\\$\downarrow$\\BUSI}} & 17.80 & 70.00 & 74.78 & \underline{77.11} & \underline{66.45} & \multirow{5}{*}{\shortstack{UCLM\\$\downarrow$\\UDIAT}} & 20.90 & \underline{63.75} & 68.15 & 82.22 & 72.06 \\

TransColor \cite{tanrscolour} & & 15.86 & 82.50 & 91.21 & 81.67 & 71.79 & & 14.21 & 72.50 & 77.28 & 75.86 & 65.38 & & 17.28 & 62.50 & \underline{68.36} & \underline{82.64} & \underline{72.56} \\

S2WAT \cite{s2wat} & & \underline{13.81} & 85.00 & 91.35 & 80.86 & 70.60 & & \underline{12.56} & 72.50 & 75.52 & 76.22 & 65.94 & & \textbf{13.04} & 61.25 & 61.12 & 80.51 & 69.98 \\

Mamba-ST \cite{mamba-st} & & 16.85 & 80.00 & 90.67 & \underline{82.69} & \underline{72.48} & & 13.36 & \underline{75.00} & \underline{78.23} & 75.19 & 64.81 & & 16.25 & 60.00 & 65.04 & 82.42 & 72.29 \\

UI-Styler & & \textbf{9.60} & \textbf{88.75} & \textbf{94.93} & \textbf{82.79} & \textbf{72.65} & & \textbf{12.40} & \textbf{80.00} & \textbf{85.60} & \textbf{80.22} & \textbf{69.78} & & \underline{13.56} & \textbf{71.25} & \textbf{73.36} & \textbf{83.16} & \textbf{73.27} \\

\midrule
w/o ST & \multirow{5}{*}{\shortstack{UDIAT\\$\downarrow$\\BUSBRA}} & 12.78 & \underline{83.67} & \underline{77.35} & \underline{87.85} & \underline{79.43} & \multirow{5}{*}{\shortstack{UDIAT\\$\downarrow$\\BUSI}} & 5.77 & 85.71 & 91.88 & \underline{84.28} & \underline{74.76} & \multirow{5}{*}{\shortstack{UDIAT\\$\downarrow$\\UCLM}} & 21.87 & 75.51 & \underline{77.14} & 85.06 & 75.60 \\

TransColor \cite{tanrscolour} & & 11.10 & 81.63 & 71.58 & 87.43 & 79.07 & & 5.68 & 83.67 & 92.09 & 83.42 & 73.98 & & 20.26 & \underline{77.55} & 73.93 & \underline{85.66} & \underline{76.19} \\

S2WAT \cite{s2wat} & & \underline{6.81} & \underline{83.67} & 74.57 & 87.63 & 79.12 & & 5.01 & 85.71 & \underline{93.38} & 81.80 & 72.10 & & \underline{17.80} & 75.51 & 71.58 & 84.55 & 75.20 \\

Mamba-ST \cite{mamba-st} & & 9.25 & 77.55 & 71.37 & 87.59 & 79.19 & & \textbf{4.38} & \underline{89.80} & 86.97 & 81.12 & 71.50 & & 18.35 & 71.43 & 75.21 & 84.35 & 74.87 \\

UI-Styler & & \textbf{5.25} & \textbf{87.76} & \textbf{79.27} & \textbf{88.45} & \textbf{80.13} & & \underline{4.47} & \textbf{91.84} & \textbf{96.15} & \textbf{85.39} & \textbf{76.09} & & \textbf{12.33} & \textbf{85.71} & \textbf{88.25} & \textbf{85.83} & \textbf{76.46} \\

\bottomrule
\end{tabular}
}
\vspace{-.7em}
\caption{
\textbf{Quantitative Comparisons.}
We evaluate the performance of unpaired image translation methods across $12$ cross-device tasks. Each group of columns corresponds to a specific source-to-target translation task. We report $5$ evaluation metrics grouped into $3$ categories:
(1) \underline{Distribution distance} — Kernel Inception Distance (KID~\textcolor{red}{$\downarrow$});
(2) \underline{Classification} — accuracy (Acc~\textcolor{red}{$\uparrow$}) and area under the ROC curve (AUC~\textcolor{red}{$\uparrow$});
(3) \underline{Segmentation} — Dice score (Dice~\textcolor{red}{$\uparrow$}) and Intersection over Union (IoU~\textcolor{red}{$\uparrow$}).
Arrows indicate whether higher or lower values are better.
The best results are shown in \textbf{bold}, while the second-best are marked with \underline{underline}. ``w/o ST" denotes without style transfer.
}
\vspace{-1em}
\label{tab:compare}
\end{table*}

\begin{figure*}[!t]
\centering
\includegraphics[width=\textwidth]{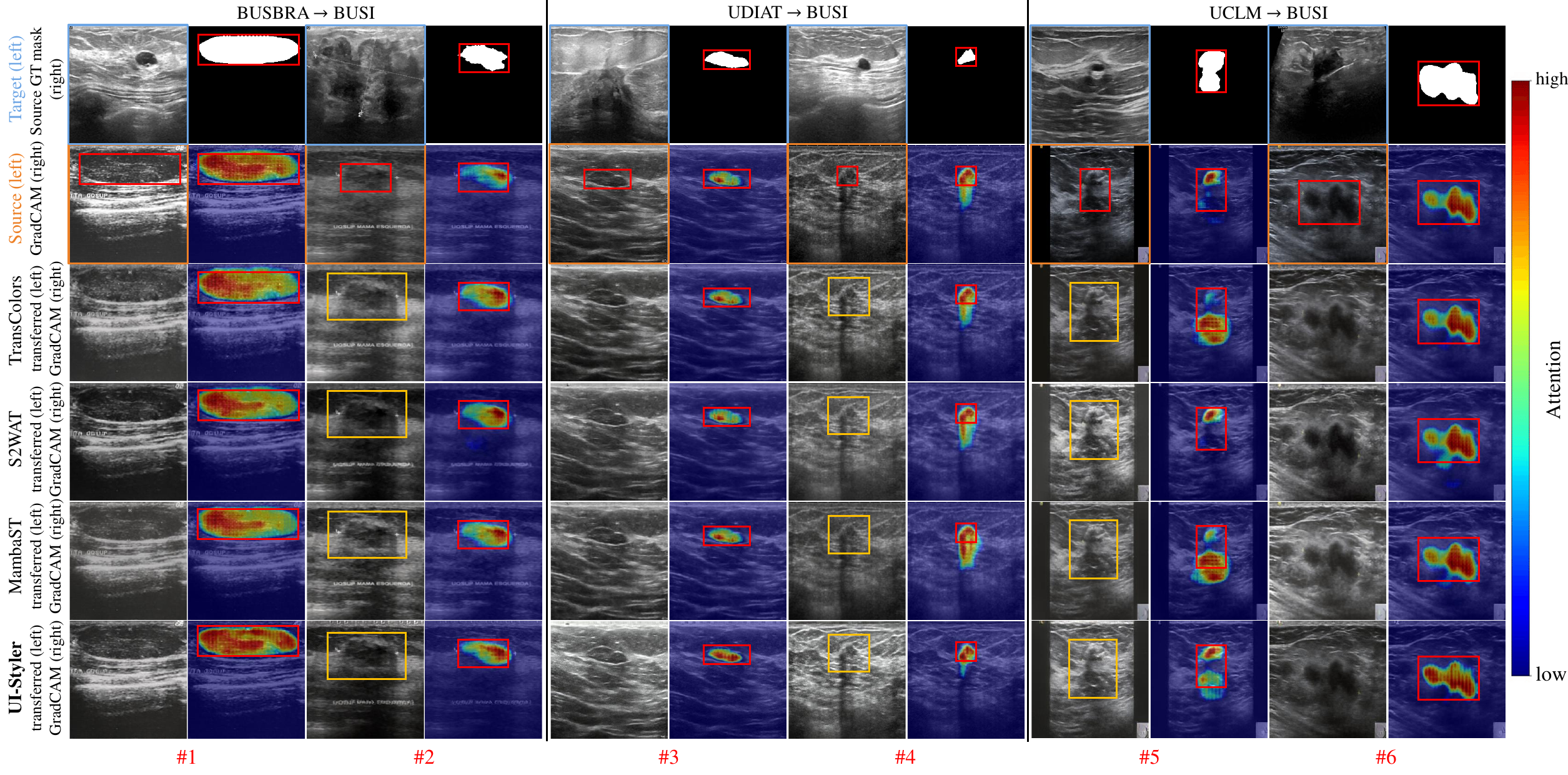}
\vspace{-2em}
\caption{
\textbf{Qualitative Comparisons.} 
We visualize Grad-CAM~\cite{gradcam} attention maps from the black-box downstream model (offline analysis only) on the BUSBRA$\rightarrow$BUSI, UDIAT$\rightarrow$BUSI, and UCLM$\rightarrow$BUSI tasks. The style reference images from the target domain are shown in the first-left row, while the source's ground-truth masks (first-right) serve as the reference for ideal attention. Each row displays the transferred images alongside the corresponding attention maps (highlighted by red squares \textcolor{red}{$\square$}) produced by different unpaired style transfer methods. Yellow squares \textcolor{DarkYellow}{$\square$} indicate regions of interest (tumor) for stylization comparison. \textit{Please zoom in to view details more easily}.
}
\label{fig:qualitative}
\vspace{-1.5em}
\end{figure*}

\section{Experiments}
\vspace{-.25em}
\subsection{Experimental Setup}
\vspace{-.25em}

\noindent \textbf{Datasets.} 
We conduct experiments on four publicly available ultrasound datasets: BUSBRA~\cite{BUS-BRA}, BUSI~\cite{BUSI}, UCLM~\cite{UCLM}, and UDIAT~\cite{UDIAT}. All datasets provide binary labels (benign vs. malignant) but differ in their acquisition devices. The number of images per dataset is listed in \cref{tab:dataset_sizes}. To simulate domain shifts, we construct $12$ transfer tasks, where each task designates one dataset as the \textit{source domain} and another as the \textit{target domain}. Each dataset is randomly split into $70\%$ training and $30\%$ testing subsets. During training, the style transfer networks are optimized using only the training subsets of both domains. At inference time, source test images are translated using style patterns from the target training set, producing stylized images that are then used for \textit{target downstream evaluation}.

\noindent \textbf{Implementation Details.}
All modules are implemented in PyTorch~\cite{PyTorch} and trained end-to-end on a single NVIDIA RTX~4090 GPU. Input images are resized to $256 \times 256$ and divided into non-overlapping patches of size $P = 8$, resulting in $L = 1024$ tokens per image. The source encoder $E_s$, target encoder $E_t$, and pattern-matching mechanism are implemented using $3$ ViT blocks~\cite{ViT}, each with an embedding dimension of $d = 512$.
All learnable parameters are initialized using Xavier initialization~\cite{glorot2010understanding}. Training is performed using the Adam optimizer \cite{Adam} with a learning rate of $5 \times 10^{-4}$, following the warm-up strategy \cite{warmlr}, a batch size of $8$, and a total of $50{,}000$ iterations.


\noindent \textbf{Evaluation Metrics.}
To quantitatively evaluate style transfer performance, we use metrics at both the distribution and task levels. 
\underline{At the distribution level}, we use the \textit{Kernel Inception Distance} (KID)~\cite{KID} to measure the distributional similarity between translated source images and target images, since it is well-suited for evaluation with small sample sizes.
\underline{At the downstream task level}, we build a \textit{black-box downstream model} (including classification and segmentation tasks) on the \textbf{target domain}'s training set.
The best-performing checkpoint is selected based on performance evaluated on the target test set and subsequently used to evaluate the translated source test images.
For the classification, we employ a ViT-B/16~\cite{ViT} model trained on images resized to $256 \times 256$ and randomly cropped to $224 \times 224$, a usable augmentation for ultrasound imaging~\cite{ultrasound_augment}. The model is optimized using stochastic gradient descent (SGD) with a learning rate of $0.001$, momentum of $0.9$, weight decay of $0.0005$, and a batch size of $16$. We report \textit{accuracy} (Acc) and \textit{area under the ROC curve} (AUC) as evaluation metrics. 
For the segmentation, we adopt SAMUS~\cite{SAMUS}, a state-of-the-art ultrasound segmentation framework, using its original training configuration. Evaluation metrics include the \textit{Dice score} and \textit{intersection over union} (IoU).
\textit{We provide performance of the black-box downstream model on target domains in Sec.~\textcolor{wacvblue}{E} of the supplementary material}.

\begin{table*}[!t]
\centering
\resizebox{\textwidth}{!}{%
\begin{tabular}{
  c c||@{\hspace{.1em}}c@{\hspace{.1em}}|M M M M M|
  @{\hspace{.1em}}c@{\hspace{.1em}}|M M M M M|
  @{\hspace{.1em}}c@{\hspace{.1em}}|M M M M M@{\hspace{.1em}}}
\toprule
\textbf{PM} & \textbf{CP} & 
\textbf{Tasks} & 
KID\textcolor{red}{$\downarrow$} & Acc\textcolor{red}{$\uparrow$} & AUC\textcolor{red}{$\uparrow$} & Dice\textcolor{red}{$\uparrow$} & IoU\textcolor{red}{$\uparrow$} &
\textbf{Tasks} & 
KID\textcolor{red}{$\downarrow$} & Acc\textcolor{red}{$\uparrow$} & AUC\textcolor{red}{$\uparrow$} & Dice\textcolor{red}{$\uparrow$} & IoU\textcolor{red}{$\uparrow$} &
\textbf{Tasks} & 
KID\textcolor{red}{$\downarrow$} & Acc\textcolor{red}{$\uparrow$} & AUC\textcolor{red}{$\uparrow$} & Dice\textcolor{red}{$\uparrow$} & IoU\textcolor{red}{$\uparrow$} \\
\midrule

\multicolumn{2}{c||@{\hspace{.1em}}}{w/o ST} & \multirow{3.2}{*}{\shortstack{BUSBRA\\$\downarrow$\\BUSI}} & 17.74 & 71.40 & 73.35 & \underline{83.99} & \underline{74.05} & \multirow{3.2}{*}{\shortstack{BUSBRA\\$\downarrow$\\UCLM}} & 28.48 & \underline{64.12} & \underline{67.80} & 81.66 & 71.01 & \multirow{3.2}{*}{\shortstack{BUSBRA\\$\downarrow$\\UDIAT}} & 13.81 & 55.95 & 64.29 & 84.76 & 75.71 \\

\textcolor{red}{\checkmark} & -- & & \underline{13.88} & \underline{72.82} & \underline{74.12} & 83.86 & 74.00 & & \underline{19.24} & 63.77 & 65.99 & \textbf{82.11} & \underline{71.65} & & \underline{12.01} & \underline{65.36} & \underline{68.29} & \underline{85.76} & \underline{76.81} \\

\textcolor{red}{\checkmark} & \textcolor{red}{\checkmark} & & \textbf{11.20} & \textbf{75.84} & \textbf{76.33} & \textbf{84.52} & \textbf{74.74} & & \textbf{16.91} & \textbf{75.13} & \textbf{76.78} & \underline{82.06} & \textbf{71.73} & & \textbf{9.14} & \textbf{72.47} & \textbf{71.52} & \textbf{86.04} & \textbf{77.52} \\

\midrule

\multicolumn{2}{c||@{\hspace{.1em}}}{w/o ST} & \multirow{3.2}{*}{\shortstack{BUSI\\$\downarrow$\\BUSBRA}} & 19.73 & 82.56 & \underline{87.30} & 82.41 & 73.37 & \multirow{3.2}{*}{\shortstack{BUSI\\$\downarrow$\\UCLM}} & 18.39 & 65.64 & 68.77 & 77.65 & 67.97 & \multirow{3.2}{*}{\shortstack{BUSI\\$\downarrow$\\UDIAT}} & 7.23 & 73.33 & 73.16 & 79.53 & 70.61 \\

\textcolor{red}{\checkmark} & -- & & \textbf{10.87} & \underline{83.59} & 87.16 & \underline{82.97} & \underline{73.99} & & \underline{13.64} & \underline{72.82} & \underline{76.97} & \underline{78.25} & \underline{68.60} & & \underline{5.60} & \textbf{74.87} & \underline{78.59} & \underline{80.38} & \underline{71.43} \\

\textcolor{red}{\checkmark} & \textcolor{red}{\checkmark} &  & \underline{11.25} & \textbf{85.13} & \textbf{88.14} & \textbf{83.15} & \textbf{74.05} & & \textbf{11.05} & \textbf{74.36} & \textbf{77.15} & \textbf{78.83} & \textbf{68.61} & & \textbf{3.61} & \underline{74.36} & \textbf{78.89} & \textbf{80.49} & \textbf{71.61} \\

\midrule

\multicolumn{2}{c||@{\hspace{.1em}}}{w/o ST} & \multirow{3.2}{*}{\shortstack{UCLM\\$\downarrow$\\BUSBRA}} & 26.74 & \underline{87.50} & 92.29 & 81.68 & 71.73 & \multirow{3.2}{*}{\shortstack{UCLM\\$\downarrow$\\BUSI}} & 17.80 & 70.00 & 74.78 & 77.11 & 66.45 & \multirow{3.2}{*}{\shortstack{UCLM\\$\downarrow$\\UDIAT}} & 20.90 & \underline{63.75} & \underline{68.15} & 82.22 & 72.06 \\

\textcolor{red}{\checkmark} & -- & & \underline{12.21} & \underline{87.50} & \underline{92.83} & \underline{82.10} & \underline{72.08} & & \underline{14.80} & \underline{77.50} & \underline{83.77} & \underline{79.81} & \underline{69.31} & & \underline{16.22} & \underline{63.75} & 67.88 & \underline{82.93} & \underline{72.96} \\

\textcolor{red}{\checkmark} & \textcolor{red}{\checkmark} & & \textbf{9.60} & \textbf{88.75} & \textbf{94.93} & \textbf{82.79} & \textbf{72.65} & & \textbf{12.40} & \textbf{80.00} & \textbf{85.60} & \textbf{80.22} & \textbf{69.78} & & \textbf{13.56} & \textbf{71.25} & \textbf{73.36} & \textbf{83.16} & \textbf{73.27}  \\

\midrule

\multicolumn{2}{c||@{\hspace{.1em}}}{w/o ST} & \multirow{3.2}{*}{\shortstack{UDIAT\\$\downarrow$\\BUSBRA}} & 12.78 & 83.67 & \underline{77.35} & 87.85 & 79.43 & \multirow{3.2}{*}{\shortstack{UDIAT\\$\downarrow$\\BUSI}} & 5.77 & 85.71 & 91.88 & \underline{84.28} & \underline{74.76} & \multirow{3.2}{*}{\shortstack{UDIAT\\$\downarrow$\\UCLM}} & 21.87 & 75.51 & 77.14 & \underline{85.06} & 75.60 \\

\textcolor{red}{\checkmark} & -- & & \underline{7.70} & \underline{85.71} & \underline{77.35} & \underline{88.15} & \underline{79.77} & & \underline{5.30} & \underline{87.76} & \underline{92.74} & 83.28 & 73.68 & & \underline{12.39} & \underline{83.67} & \underline{87.18} & 85.01 & \underline{75.63} \\

\textcolor{red}{\checkmark} & \textcolor{red}{\checkmark} & & \textbf{5.25} & \textbf{87.76} & \textbf{79.27} & \textbf{88.45} & \textbf{80.13} & & \textbf{4.47} & \textbf{91.84} & \textbf{96.15} & \textbf{85.39} & \textbf{76.09} & & \textbf{12.33} & \textbf{85.71} & \textbf{88.25} & \textbf{85.83} & \textbf{76.46} \\
           
\bottomrule
\end{tabular}
}
\vspace{-.5em}
\caption{\textbf{Ablation Study.} We evaluate the contribution of the pattern-matching (PM) and class-aware prompting (CP) modules across $12$ cross-device ultrasound tasks with $5$ metrics: KID, Acc, AUC, Dice, and IoU. \textbf{Bold} marks the best results; \underline{underline} for second-best.}
\vspace{-.5em}
\label{tab:ablation}
\end{table*}

\begin{figure*}[!t]
\centering
\begin{subfigure}[b]{0.27\linewidth}
    \includegraphics[width=\linewidth]{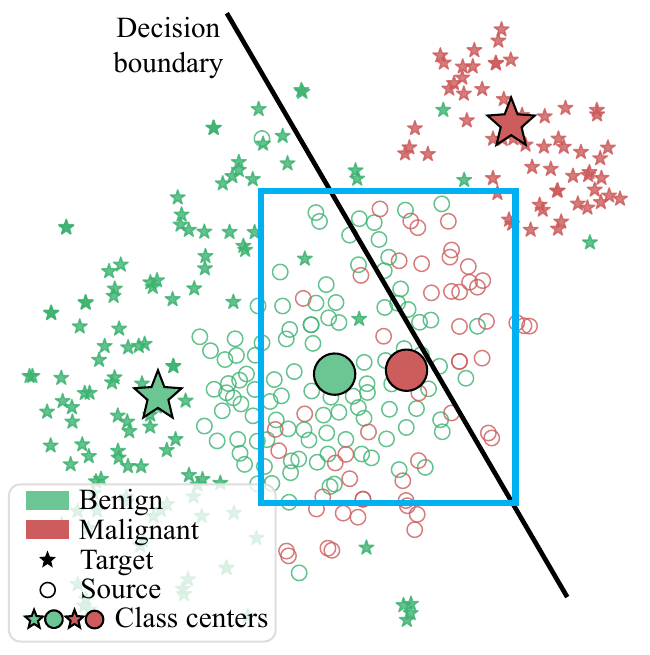}
    \subcaption{Before style transfer.}
    \label{fig:image2a}
\end{subfigure}
\hfill
\begin{subfigure}[b]{0.27\linewidth}
    \includegraphics[width=\linewidth]{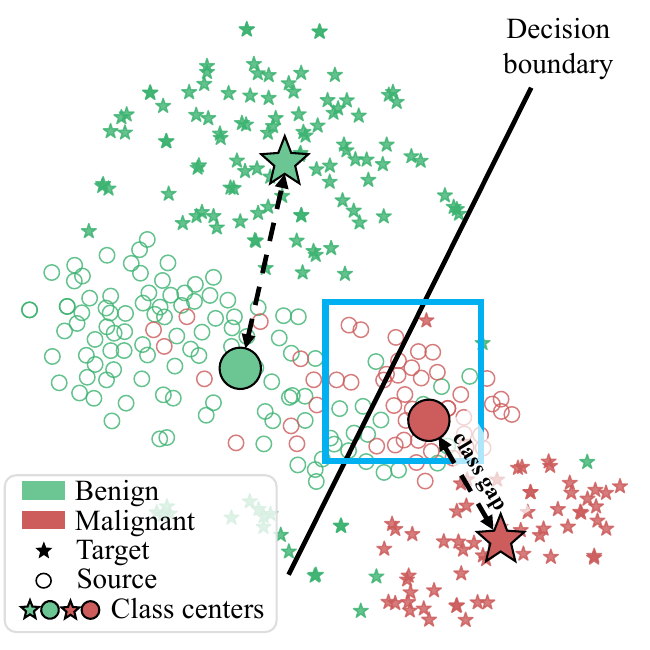}
    \subcaption{Only domain level.}
    \label{fig:image2b}
\end{subfigure}
\hfill
\begin{subfigure}[b]{0.27\linewidth}
    \includegraphics[width=\linewidth]{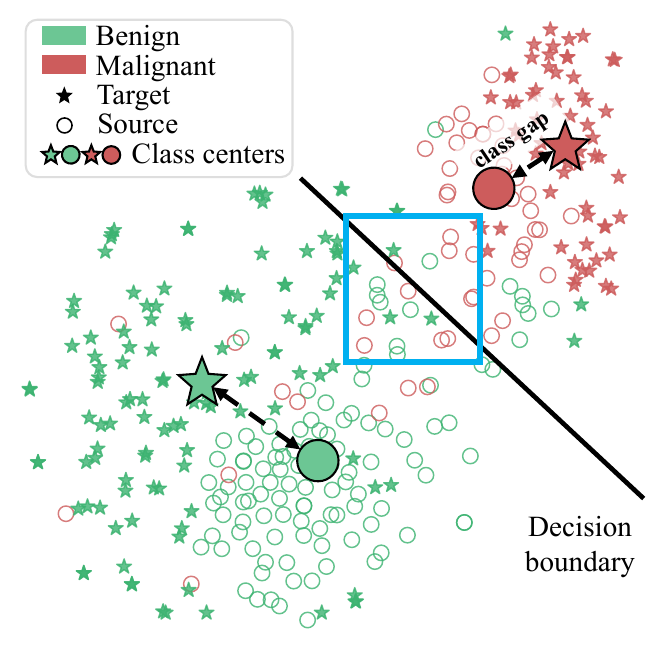}
    \subcaption{Domain and class levels.}
    \label{fig:image2c}
\end{subfigure}

\vspace{-.5em}
\caption{\textbf{Feature Space.} We visualize the feature distributions using t-SNE~\cite{t-SNE} on the UDIAT$\rightarrow$UCLM task. Each point represents a sample: \textcolor{ForestGreen}{green} for benign and \textcolor{red}{red} for malignant. $\star$ indicates target samples (UCLM), while $\circ$ denotes source samples (UDIAT) under three conditions—(a) before translation, (b) after domain-level alignment only, and (c) after full dual-level stylization by UI-Styler.}
\label{fig:tsne}
\vspace{-1.em}
\end{figure*}

\subsection{Comparison Results}

\noindent \textbf{Quantitative Comparisons.}
Table~\ref{tab:compare} reports results across $12$ cross-device ultrasound tasks using $5$ metrics spanning distribution distance (KID) and task-level performance (Acc, AUC, Dice, IoU). \textbf{UI-Styler consistently achieves top performance across all metrics}. Specifically, UI-Styler yields the lowest KID in most tasks, confirming superior distribution matching.
\underline{In classification}, UI-Styler improves accuracy by $+5.00\%$ over Mamba-ST~\cite{mamba-st} on UCLM$\rightarrow$BUSI and AUC by $+2.77\%$ over S2WAT~\cite{s2wat} on UDIAT$\rightarrow$BUSI. \underline{In segmentation}, it surpasses TransColor~\cite{tanrscolour}, a method specialized in ultrasound imaging, by $+0.52$ in Dice and $+0.71$ in IoU on UCLM$\rightarrow$UDIAT.
\textit{More broadly}, prior UIT methods tend to focus on minimizing domain-level appearance discrepancies, inadvertently leading to misalignment at the class level. As evident in BUSI$\rightarrow$UCLM and UCLM$\rightarrow$BUSBRA in terms of Acc and AUC, as well as BUSBRA$\rightarrow$BUSI and UDIAT$\rightarrow$BUSBRA in terms of Dice and IoU, where prior methods perform worse than those without style transfer (w/o ST). 
In contrast, UI-Styler's dual-level stylization effectively bridges both domain and class gaps, resulting in consistently stable and superior results.

\noindent \textbf{Qualitative Comparisons.}
To assess the impact of style translation results on downstream model behavior, we visualize Grad-CAM~\cite{gradcam} attention maps from the black-box downstream model on $3$ cross-device tasks: BUSBRA$\rightarrow$BUSI, UDIAT$\rightarrow$BUSI, and UCLM$\rightarrow$BUSI. Ideally, attention maps should exhibit high activation values localized within tumor regions, consistent with the ground-truth masks. As shown in \cref{fig:qualitative}, prior methods such as TransColor~\cite{tanrscolour}, S2WAT~\cite{s2wat}, and Mamba-ST~\cite{mamba-st} often produce incomplete attention (e.g., columns $\#1$, $\#3$, $\#6$) or noisy, redundant activations (e.g., columns $\#4$, $\#5$), highlighted in \textcolor{red}{red} squares. Moreover, we use the \textcolor{DarkYellow}{yellow} squares to highlight regions of interest for comparison. In prior works, some translated images exhibit blurred lesion boundaries (e.g., column~$\#2$) or fail to distinguish between tumor and non-tumor regions (e.g., column~$\#4$).
In contrast, UI-Styler generates attention maps that align more closely with the ground-truth masks. Even in challenging cases where the source visual contrast is low, UI-Styler achieves clear tumor delineation and reliable attention, thereby facilitating accurate segmentation.

\subsection{Analysis} \label{exp:analysis}

\begin{figure*}[ht]
\centering
\begin{subfigure}[b]{0.21\textwidth}
    \includegraphics[width=\linewidth]{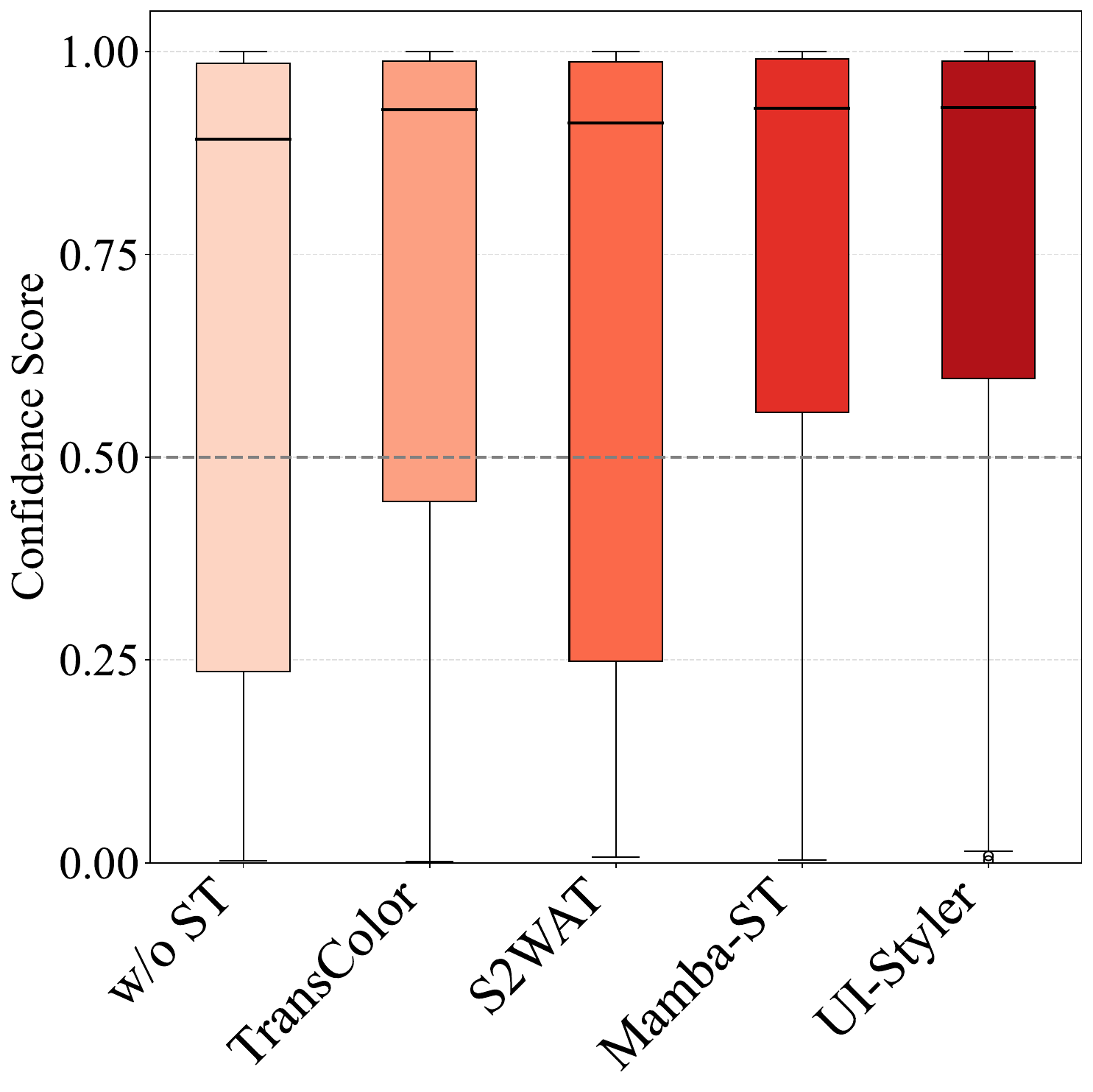}
    \caption{BUSBRA$\rightarrow$BUSI.}
    \label{fig:box1}
\end{subfigure}
\hfill
\begin{subfigure}[b]{0.21\textwidth}
    \includegraphics[width=\linewidth]{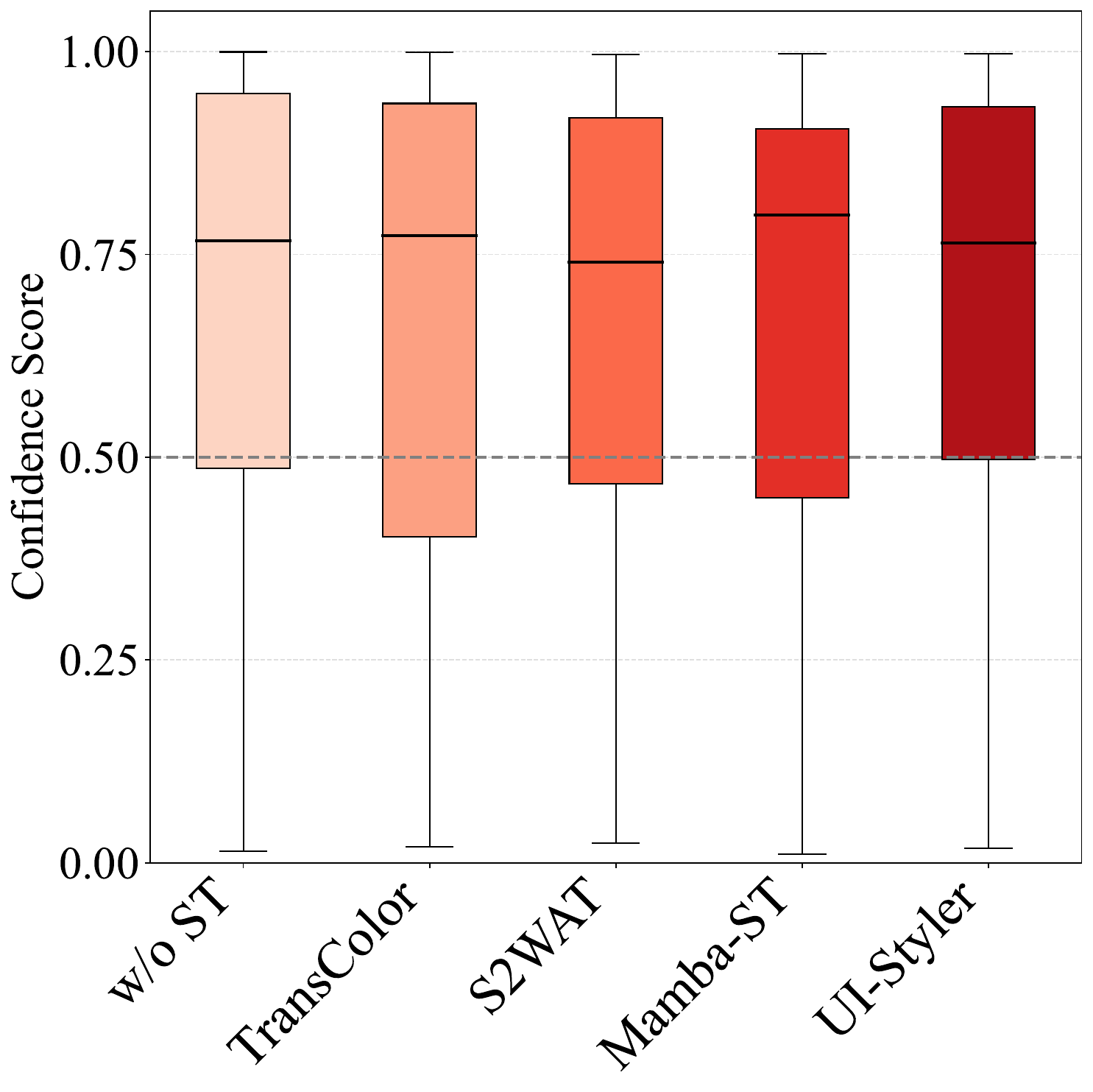}
    \caption{BUSI$\rightarrow$UDIAT.}
    \label{fig:box2}
\end{subfigure}
\hfill
\begin{subfigure}[b]{0.21\textwidth}
    \includegraphics[width=\linewidth]{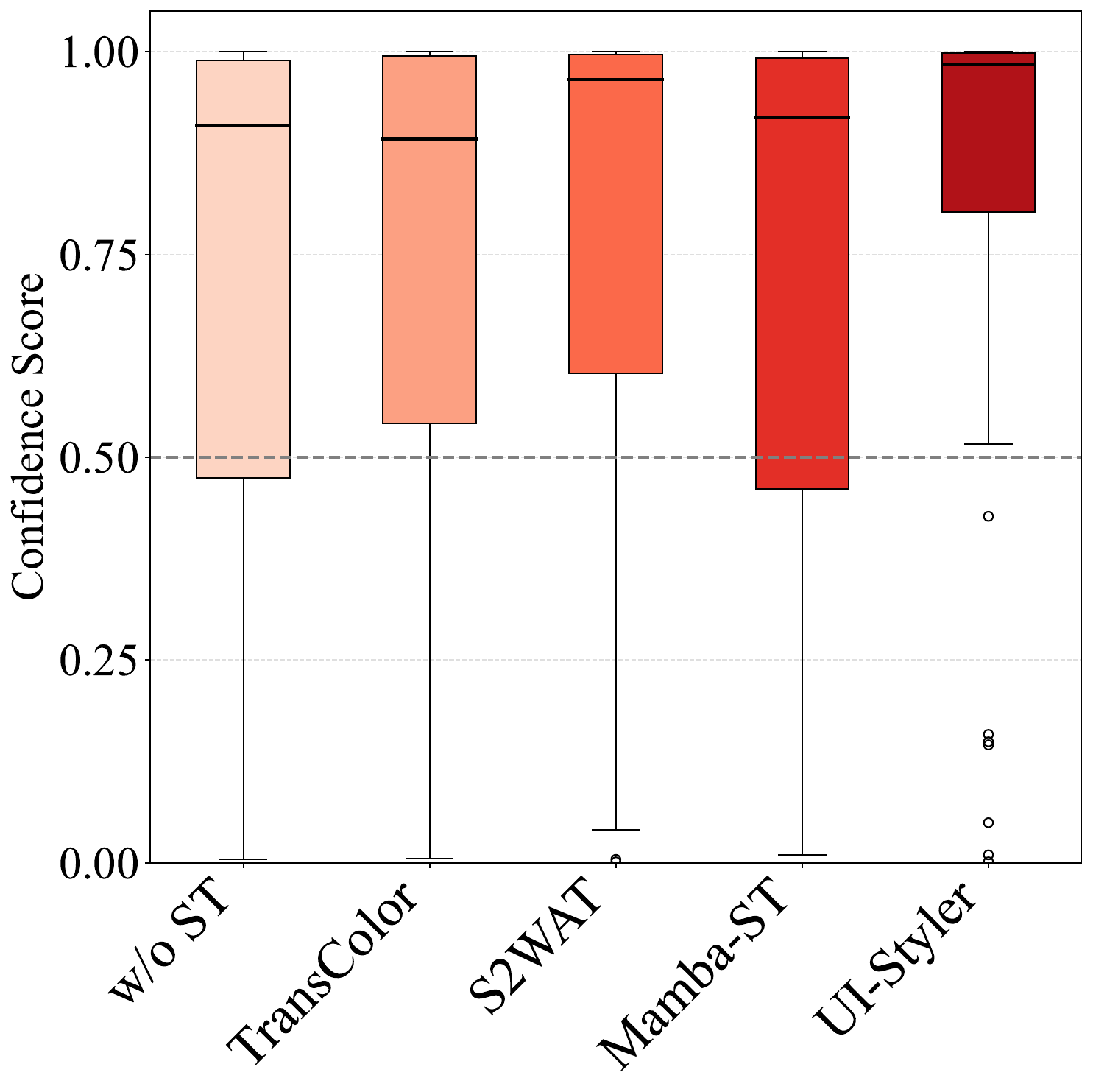}
    \caption{UDIAT$\rightarrow$UCLM.}
    \label{fig:box3}
\end{subfigure}
\hfill
\begin{subfigure}[b]{0.21\textwidth}
    \includegraphics[width=\linewidth]{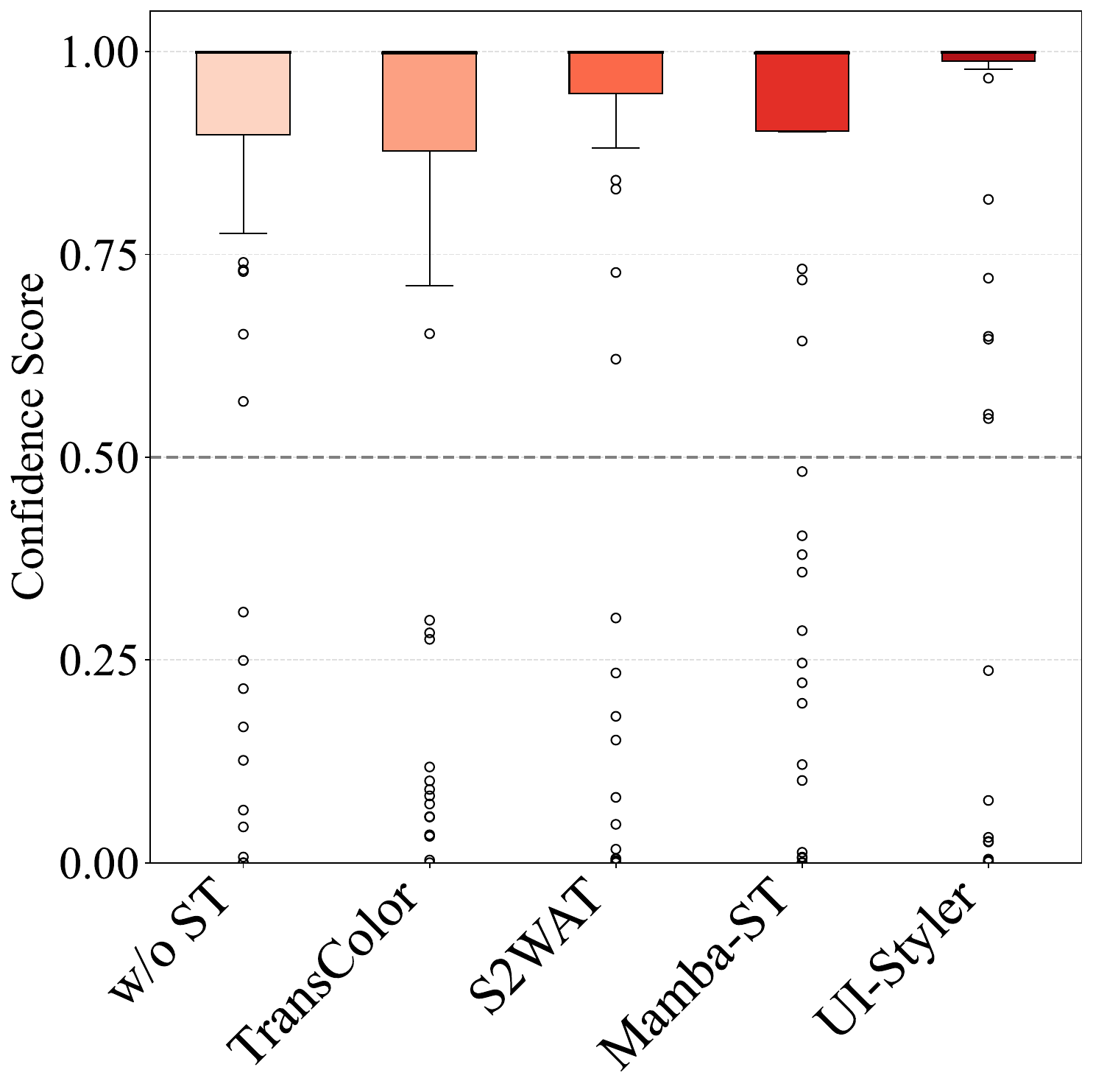}
    \caption{UCLM$\rightarrow$BUSBRA.}
    \label{fig:box4}
\end{subfigure}
\hfill
\begin{subfigure}[b]{0.141\textwidth}
    \includegraphics[width=\linewidth]{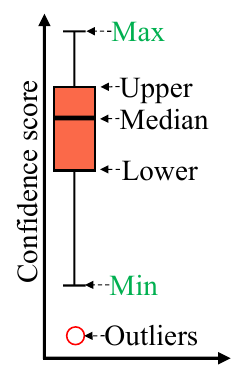}
    \caption{Boxplot.}
    \label{fig:box5}
\end{subfigure}

\vspace{-.7em}
\caption{%
\textbf{Confidence Scores.} We visualize the distribution of confidence scores predicted by the black-box downstream model on stylized-source test samples across $4$ source-to-target adaptation tasks. Each box plot shows the predicted probability assigned to the ground-truth class.
(e) In the boxplot, the median indicates central prediction confidence, the box spans the interquartile range, and the min–max lines show the full prediction spread. Outliers highlight irregular cases. Higher medians and tighter boxes indicate more confident predictions.
} \label{fig:confidence}
\vspace{-1em}
\end{figure*}

\noindent \textbf{Ablation Study.}
To evaluate the effectiveness of each component in UI-Styler, we perform an ablation study assessing the impact of the pattern-matching module (PM) and the class-aware prompting module (CP) across multiple cross-device ultrasound tasks, as reported in Table~\ref{tab:ablation}.
The \textit{pattern-matching module} serves as the foundation for domain-level adaptation by aligning source content with target style. When enabled alone (PM only), it substantially reduces KID and improves both classification and segmentation performance compared to the no-style-transfer baseline (w/o ST). For example, on BUSI$\rightarrow$UCLM, PM lowers KID from $18.39$ to $13.64$ and boosts AUC by $+8.20$\%.
Building on this, the \textit{class-aware prompting module} further enhances the semantic alignment of the stylized features produced by PM. When CP is added (i.e., full UI-Styler), we observe consistent improvements across nearly all evaluation metrics. For example, on UCLM$\rightarrow$UDIAT, the full configuration increases accuracy from $63.75$ to $71.25$ and improves Dice from $82.93$ to $83.16$.
These findings confirm that PM and CP jointly implement a \textbf{coarse-to-fine alignment strategy}, ensuring both domain-level appearance consistency and class-specific semantic refinement.

\noindent \textbf{Feature Space.}
We visualize the feature distributions of the black-box downstream model using t-SNE~\cite{t-SNE} on the UDIAT$\rightarrow$UCLM task in \cref{fig:tsne}. Each plot shows the 2D projection of source and target features under three configurations: (a) no style transfer, (b) domain-level stylization only, and (c) dual-level stylization with UI-Styler.
In \cref{fig:image2a}, without any adaptation, benign and malignant source features exhibit significant overlap and cannot be reliably classified, particularly in the region highlighted by the \textcolor{cyan}{blue} square.
In \cref{fig:image2b}, applying only domain-level stylization via pattern-matching reduces the domain gap. However, class-level information is not considered; source features still cluster ambiguously near the decision boundary (within the \textcolor{cyan}{blue} square) and remain far from the target class centers (indicated by the dashed lines).
In contrast, \cref{fig:image2c} shows that UI-Styler’s dual-level stylization effectively reduces both domain and class gaps. By injecting class-specific prompts, UI-Styler explicitly steers source features toward the correct side of the decision boundary. As highlighted by the \textcolor{cyan}{blue} square, this reduces inter-class confusion near the boundary and improves alignment between same-class samples (e.g., benign \textcolor{red}{\ding{109}} aligned with benign \textcolor{red}{\ding{72}}).

\noindent \textbf{Confidence Score.}
Figure~\ref{fig:confidence} shows box plots of confidence scores produced by the black-box downstream model on stylized-source test samples generated by various unpaired image translation methods. Each plot summarizes the predictive certainty under a specific source-to-target adaptation scenario. Confidence scores are computed by extracting the predicted probability corresponding to the \textit{ground-truth label}—e.g., if the ground truth is class 0 and the predicted probability for class 0 is $0.3$, the recorded score is $0.3$ regardless of the final prediction. Across all tasks, UI-Styler consistently achieves a higher median confidence and a narrow interquartile range, reflecting strong semantic preservation. While Mamba-ST~\cite{mamba-st} shows competitive performance in certain tasks (e.g., BUSBRA$\rightarrow$BUSI), it suffers from higher variance
than UI-Styler. TransColor~\cite{tanrscolour} and S2WAT~\cite{s2wat} display broader distributions with lower medians, making some scores fall below the $0.5$ decision threshold, especially in challenging scenarios such as UDIAT$\rightarrow$UCLM and UCLM$\rightarrow$BUSBRA.
These observations underscore a key limitation of prior methods: although transferring style, they often fail to preserve class-specific characteristics. In contrast, UI-Styler leads to improvements in both accuracy and confidence robustness.

\vspace{-.5em}
\section{Conclusion}
\vspace{-.5em}

In this work, we propose \textbf{UI-Styler}, a novel, ultrasound-specific, class-aware framework for unpaired image translation under an inference-blackbox setting. Unlike prior approaches that focus solely on minimizing the domain-level style discrepancies, UI-Styler introduces a \textit{dual-level stylization module}—combining a pattern-matching mechanism with class-aware prompting—to achieve both domain-level and class-level alignment. Our method is trained without requiring access to source or target labels, logits, or backbone gradients, making it particularly suitable for privacy-sensitive and label-scarce medical scenarios. Extensive experiments on $12$ cross-device ultrasound tasks demonstrate that UI-Styler outperforms existing unpaired image translation methods in terms of distribution alignment as well as downstream tasks, such as classification and segmentation.


\section*{Acknowledgments}
This work was financially supported in part (project number: 112UA10019) by the Co-creation Platform of the Industry Academia Innovation School, NYCU, under the framework of the National Key Fields Industry-University Cooperation and Skilled Personnel Training Act, from the Ministry of Education (MOE) and industry partners in Taiwan.  It also supported in part by the National Science and Technology Council, Taiwan, under Grant NSTC-112-2221-E-A49-089-MY3, Grant NSTC-110-2221-E-A49-066-MY3, Grant NSTC-111-2634-F-A49-010, Grant NSTC-112-2425-H-A49-001, and in part by the Higher Education Sprout Project of the National Yang Ming Chiao Tung University and the Ministry of Education (MOE), Taiwan. We also would like to express our gratitude for the support from MediaTek Inc, Hon Hai Research Institute (HHRI),  E.SUN Financial Holding Co Ltd, Advantech Co Ltd, Industrial Technology Research Institute (ITRI).

{
    \small
    \bibliographystyle{ieeenat_fullname}
    \bibliography{main}
}

\end{document}


\twocolumn[{%
\renewcommand\twocolumn[1][]{#1}%
\maketitlesupplementary
\begin{center}
    \captionsetup{type=figure}
    \begin{subfigure}[b]{0.23\textwidth}
        \centering
        \includegraphics[width=.87\linewidth]{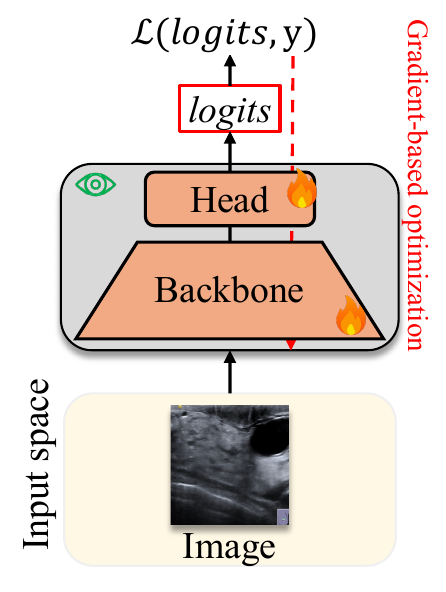}
        \subcaption{Full white-box setting.}
        \label{fig:full}
    \end{subfigure}
    \hfill
    \begin{subfigure}[b]{0.23\textwidth}
        \centering
        \includegraphics[width=.87\linewidth]{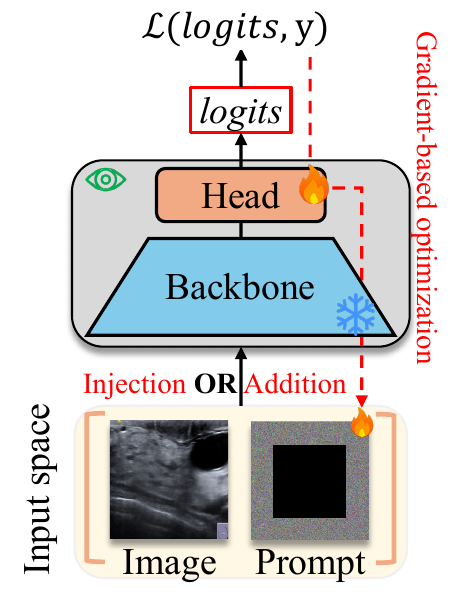}
        \subcaption{Prompt white-box setting.}
        \label{fig:prompt}
    \end{subfigure}
    \hfill
    \begin{subfigure}[b]{0.23\textwidth}
        \centering
        \includegraphics[width=\linewidth]{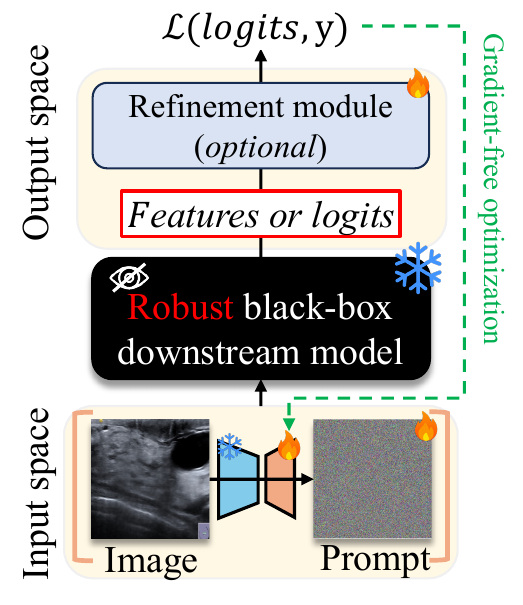}
        \subcaption{Black-box setting.}
        \label{fig:blackbox}
    \end{subfigure}
    \hfill
    \begin{subfigure}[b]{0.23\textwidth}
        \centering
        \includegraphics[width=0.95\linewidth]{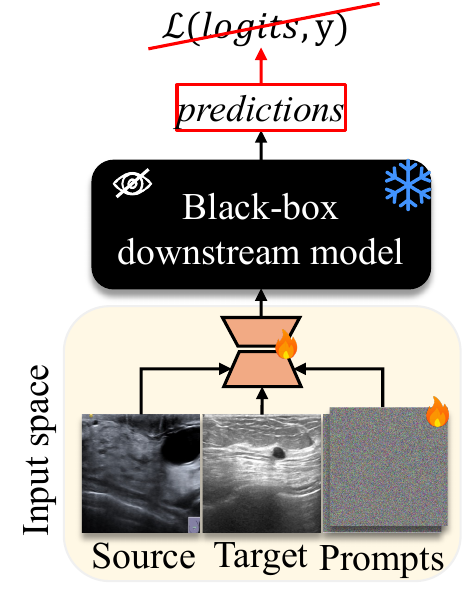}
        \subcaption{Inference-blackbox setting.}
        \label{fig:inference}
    \end{subfigure}
    \setcounter{figure}{0}
    \vspace{-.5em}
    \captionof{figure}{
    \textbf{Prompt Setting Comparison.} We illustrate four prompt-based training and deployment scenarios with increasing constraints: 
    (a) \textit{Full white-box setting} allows end-to-end fine-tuning via backpropagation over the entire model using ground-truth labels. 
    (b) \textit{Prompt white-box setting} injects learnable prompts into the input while freezing the backbone, but still requires gradients and supervision. 
    (c) \textit{Black-box setting} removes gradient access but assumes availability of intermediate features or logits for prompt tuning or refinement. 
    (d) \textit{Inference-blackbox setting} reflects the most realistic and constrained scenario, where only final predictions are available.
    }
    \label{fig:setting}
\end{center}%
}]

\section*{Overview}
\vspace{-.5em}

We organize the supplementary content into nine sections. 
\cref{sec:notation} introduces key notations, and \cref{sec:code} provides the pseudo-code of UI-Styler. 
\cref{sec:setting} compares full fine-tuning and prompt-tuning paradigms under different levels of model access, while \cref{sec:loss} details the content and style losses. 
\cref{sec:target_perf} reports black-box downstream performance on target domains. 
\cref{sec:extra_exp} presents additional experiments on loss contributions, weight configurations, and pattern-matching sensitivity. 
\cref{sec:extra_analysis} further analyzes diagnostic semantic preservation and t-SNE failure cases. 
\cref{sec:discussion} discusses scalability, generalization, and robustness to noisy pseudo labels. 
Finally, \cref{sec:visual} provides qualitative results across all $12$ cross-device tasks.

\vspace{-1.em}

\begingroup
\small 
\tableofcontents
\endgroup

\vspace{-.5em}
\section{Notation} \label{sec:notation}
\vspace{-.5em}
We summarize the notations and their corresponding definitions frequently used in our method in~\cref{tab:notation}.

\vspace{-.5em}
\section{Pseudo Code} \label{sec:code}
\vspace{-.5em}
We provide the pseudo code of \textbf{UI-Styler} in Algorithm~\ref{algorithm_1}, which outlines the core procedures for training and testing. 


\vspace{-.5em}
\section{Problem Setting Comparison} \label{sec:setting}
\vspace{-.5em}

In this section, we categorize and compare four increasingly constrained training and deployment scenarios, ranging from full fine-tuning in white-box settings to prompt tuning under inference-blackbox conditions. Each setting imposes distinct assumptions on parameter accessibility, label availability, and interaction scope, as summarized in~\cref{fig:setting}. We highlight the practical limitations of existing methods in real-world deployment scenarios, thus motivating our inference-blackbox prompt tuning.

\begin{table}[H]
\centering
\renewcommand{\arraystretch}{1.1}
\resizebox{\columnwidth}{!}{\Large
\begin{tabular}{@{\hspace{.1em}}l@{\hspace{.1em}}|@{\hspace{.1em}}l@{\hspace{.1em}}}
\toprule
\LARGE \textcolor{blue}{Symbol} & \multicolumn{1}{c}{\LARGE \textcolor{blue}{Description}} \\
\midrule
\multicolumn{2}{c}{\textbf{Abbreviations}} \\
\midrule
BDM & Black-box downstream model \\
PT & Prompt tuning \\
PDA & Prompt-based domain adaptation \\
UIT & Unpaired image translation \\
PM & Pattern-matching mechanism (domain-level adaptation) \\
CP & Class-aware prompting (class-level alignment) \\
ViT & Vision transformer \\
\midrule
\multicolumn{2}{c}{\textbf{Data Setting}} \\
\midrule
$\mathcal{D}_s$ & Unlabeled source domain \\
$\mathcal{D}_t$ & Unlabeled target domain \\
$x_s$, $x_t$ & Source and target images \\
$\hat{y}_t$ & Pseudo target label \\
$\mathbf{\hat{y}}_t$ & One-hot encoding of the pseudo target label \\
$C$ & Number of classes  \\
$H \times W$ & Input image size ($256 \times 256$) \\
\midrule
\multicolumn{2}{c}{\textbf{UI-Styler Architecture}} \\
\midrule
$P$ & Patch size (set to $8$) \\
$h$, $w$ & Patch grid size, $h = H/P$, $w = W/P$ \\
$L$ & Number of image tokens ($L = h\times w$) \\
$d$ & Embedding dimension of each token \\
$E_s$, $E_t$ & source and target encoders \\
$W_q$ & Projection matrix for query from source features \\
$W_k$, $W_v$ & Projection matrices for key and value from target features \\
$\mathcal{E}_f(\cdot)$, $\mathcal{E}_p(\cdot)$ & Feature and prompt embedders \\
$H(\cdot)$ & A classifier head \\
$D$ & Decoder to reconstruct stylized images \\
$\widetilde{x}_s$ & Stylized image \\
\midrule
\multicolumn{2}{c}{\textbf{Features \& Representations}} \\
\midrule
$\mathcal{F}_s$, $\mathcal{F}_t$ & Extracted features from source and target images \\
$Q$ & Query, projected from $\mathcal{F}_s$ using $W_q$ \\
$K$, $V$ & Key and Value, projected from $\mathcal{F}_t$ using $W_k$, $W_v$ \\
$\widetilde{\mathcal{F}}_{s \rightarrow t}$ & Stylized features (after domain-level alignment) \\
$\widetilde{\mathcal{F}}^+_{s \rightarrow t}$ & Final stylized features (after class-aware prompting) \\
$\mathcal{P}$ & Learnable template prompts \\
$\mathcal{P}_c$ & Class-specific prompts \\
$\hat{\mathcal{P}}_c$ & Supervised prompts derived from the pseudo target label \\

\midrule
\multicolumn{2}{c}{\textbf{Loss Functions}} \\
\midrule
$\mathbf{a}$ & Class–prompt correlation vector \\
$\mathbf{p}_t$ & Probabilities from classifier head $H(\mathcal{F}_t + \hat{\mathcal{P}_c})$ \\
$\mathcal{L}_c$ & Content loss (structure/content preservation) \\
$\mathcal{L}_s$ & Style loss (appearance/style alignment) \\
$\mathcal{L}_{\text{dir}}$ & Direction loss for prompt selection \\
$\mathcal{L}_{\text{sup}}$ & Supervised loss for prompt supervision \\
$\mathcal{L}_{\text{total}}$ & Overall training objective \\
\midrule
\multicolumn{2}{c}{\textbf{Evaluation Metrics}} \\
\midrule
KID \textcolor{red}{$\downarrow$} & Kernel Inception Distance \\
Acc \textcolor{red}{$\uparrow$} & Classification accuracy \\
AUC \textcolor{red}{$\uparrow$}  & Area under ROC curve \\
Dice \textcolor{red}{$\uparrow$}  & Dice score \\
IoU \textcolor{red}{$\uparrow$}  & Intersection over Union \\
\bottomrule
\end{tabular}
}
\vspace{-.5em}
\caption{Summary of notations used throughout the paper.} \label{tab:notation}
\end{table}

\begin{algorithm}[H]
\linespread{0.9}\selectfont
    \caption{The pseudo code of UI-Styler}  \label{algorithm_1}
    \begin{algorithmic}[1] 
    
    \STATE \textbf{Problem Setting} (Sec.~\textcolor{wacvblue}{3.1}):

    \textcolor{teal!70!black}{\ding{114} Data Setting:}
    \begin{itemize}
        \item The unlabeled source dataset  $\mathcal{D}_s = \{x_s^i\}_{i=1}^{N_s}$.
        \item The pseudo-labeled target dataset $\mathcal{D}_t = \{(x_t^j, \hat{y}_t^j)\}_{j=1}^{N_t}$.
    \end{itemize}
    
    \noindent \textcolor{red}{Note}: Unpaired source and target data,  $\mathcal{D}_s \cap \mathcal{D}_t = \emptyset$. \vspace{0.5em}

    \textcolor{teal!70!black}{\ding{114} Black-box Downstream Model:} classification network: $C(\cdot)$ and segmentation network: $S(\cdot)$.

    \STATE \textbf{UI-Styler Architecture} (Sec.~\textcolor{wacvblue}{3.2}):

    \noindent \textcolor{teal!70!black}{\ding{114} Feature Extractors}: a source encoder $E_s(\cdot;\boldsymbol{\theta}_{E_s})$ and a target encoder $E_t(\cdot;\boldsymbol{\theta}_{E_t})$.

    \noindent \textcolor{teal!70!black}{\ding{114} Dual-level Stylization}:
    
    \begin{itemize} 
        \item \underline{Pattern-matching Mechanism}:
        
        \hspace{-1.2em} $\text{PM}(c,s; \boldsymbol{\theta}_{PM}) = \{W_q(c;\boldsymbol{\theta}_{W_q}), W_k(s;\boldsymbol{\theta}_{W_k}), W_v(s;\boldsymbol{\theta}_{W_v})\}$.
        
        \item \underline{Class-aware Prompting}:
        
        \hspace{-1.2em} $\text{CP}(\cdot,\cdot; \boldsymbol{\theta}_{CP}) = \{ \mathcal{P}(\boldsymbol{\theta}_{\mathcal{P}}), \mathcal{E}_f(\cdot;\boldsymbol{\theta}_{\mathcal{E}_f}), \mathcal{E}_p(\cdot;\boldsymbol{\theta}_{\mathcal{E}_p}), H(\cdot;\boldsymbol{\theta}_{H})\}$.
    \end{itemize}

    \noindent \textcolor{teal!70!black}{\ding{114} Decoder}: $D(\cdot; \boldsymbol{\theta}_{D})$.

    \noindent \textcolor{red}{Note}: Parameters: $\textcolor{red}{\theta}=\{\boldsymbol{\theta}_{E_s}, \boldsymbol{\theta}_{E_t}, \boldsymbol{\theta}_{PM}, \boldsymbol{\theta}_{CP}, \boldsymbol{\theta}_{D}\}$ is initialized using Xavier and optimized with learning rates $\eta$.


    \STATE \textbf{Training Strategy:}
    
    \FOR {$i {\leftarrow} 1$ to $I$}
        \STATE {\textcolor{violet}{\ding{52} Feature Extraction}} (Sec.~\textcolor{wacvblue}{3.2}):

        $\mathcal{F}_s = E_s(x^i_s), \quad
        \mathcal{F}_t = E_t(x^i_t)$,

        \STATE {\textcolor{violet}{\ding{52} Dual-level Stylization}} (Sec.~\textcolor{wacvblue}{3.3}):

        \STATE {\textcolor{orange}{{\ding{48}} 1. } \underline{Domain-level adaptation}}

        \textit{\# Stylized Features}
        
        $\widetilde{\mathcal{F}}_{s \rightarrow t}=\text{PM}(\mathcal{F}_s, \mathcal{F}_t)$, \hfill \textcolor{magenta}{{$\rhd$}} Eqs. \textcolor{wacvblue}{1, 2}.

        \STATE {\textcolor{orange}{{\ding{48}} 2.} \underline{Class-level adaptation}}

        \textit{\# Class-specific Prompts}
        
        $\mathcal{P}_c = \mathsf{\text{one-hot-max}}\left(\mathcal{E}_f(\widetilde{\mathcal{F}}_{s \rightarrow t}) \mathcal{E}_p(\mathcal{P})^\top\right) \mathcal{P}$,
        
        \hfill \textcolor{magenta}{{$\rhd$}} Eq. \textcolor{wacvblue}{3}.

        \textit{\# Class-aligned Features}
        
        $\widetilde{\mathcal{F}}^+_{s \rightarrow t} = \widetilde{\mathcal{F}}_{s \rightarrow t} + \mathcal{P}_c$, \hfill \textcolor{magenta}{{$\rhd$}} Eq. \textcolor{wacvblue}{4}.

        \STATE {\textcolor{violet}{\ding{52} Reconstruction}} (Sec.~\textcolor{wacvblue}{3.2}):        
        
        $\widetilde{x}_s = D(\widetilde{F}^+_{s \rightarrow t})$,

        \STATE {\textcolor{WildStrawberry}{\ding{224} \textbf{Final Objective Function}}} (Sec.~\textcolor{wacvblue}{3.4}):

        \textit{\# Direction Loss} \\
        $\mathbf{a} = \mathrm{sigmoid}(\mathcal{E}_f(\mathcal{F}_t) \cdot \mathcal{E}_p(\mathcal{P})^\top) \in \mathbb{R}^{C}$,

        $\mathcal{L}_{\text{dir}} = - \frac{1}{C} \sum_{c=1}^{C} \left[ \hat{y}_c \log a_c + (1 - \hat{y}_c) \log (1 - a_c) \right]$, \\        
        \hfill \textcolor{magenta}{{$\rhd$}} Eq. \textcolor{wacvblue}{5}.
        
        \textit{\# Supervised Loss} \\
        
        $\hat{\mathcal{P}_c} = \mathbf{\hat{y}}_t \cdot \mathcal{P} \in \mathbb{R}^{L \times d}$, \\
        
        $\mathcal{L}_{\text{sup}} = - \mathbf{\hat{y}}_t \cdot \log(\mathbf{p}_t)$,
        \\        
        \hfill where $\mathbf{p}_t = \mathrm{softmax}(H(\mathcal{F}_t + \hat{\mathcal{P}_c}))$ \textcolor{magenta}{{$\rhd$}} Eq. \textcolor{wacvblue}{6}.
        
        \textit{\# Backpropagation} \\
        $\mathcal{L}_{\text{total}} = \lambda_{\text{dir}}\mathcal{L}_{\text{dir}} + \lambda_{\text{sup}}\mathcal{L}_{\text{sup}} + \lambda_c\mathcal{L}_c + \lambda_s\mathcal{L}_s$, \\
        
        $\textcolor{red}{\theta} \leftarrow \textcolor{red}{\theta} - \eta \nabla_{\boldsymbol{\theta}} \mathcal{L}_{\text{total}}$.

    \ENDFOR \vspace{0.5em}

    \STATE \textbf{Testing:} \\
    \textcolor{teal!70!black}{\ding{114} Style Transfer}: $\widetilde{x}_s = \text{UI-Styler}(x_s, x_t)$, \\
    \textcolor{teal!70!black}{\ding{114} Reused Black-box Downstream Model}:
    \begin{itemize}
        \item \textcolor{orange!80!black}{Predicted Class}: $\hat{y}_{s \rightarrow t} = C(\widetilde{x}_s)$,
        \item \textcolor{orange!80!black}{Predicted Mask}: $\hat{M}_{s \rightarrow t} = S(\widetilde{x}_s)$.
    \end{itemize}
    \end{algorithmic}
\end{algorithm}

\subsection{Full Fine-Tuning in White-box Setting}
As shown in~\cref{fig:full}, full fine-tuning (FT) enables end-to-end optimization of both the backbone and task-specific head using supervised loss $\mathcal{L}(\textit{logits}, y)$, where $y$ is the ground truth. Despite achieving strong task-specific performance~\cite{finetuning, finetune2}, FT demands full access to model parameters and gradients, making it infeasible in proprietary or privacy-sensitive deployments. Moreover, it incurs high computational overhead and risks of overfitting or catastrophic forgetting under distribution shifts.

\subsection{Prompt Tuning in White-box Setting}
Prompt tuning (PT) alleviates the limitations of FT by inserting learnable prompts into the input space while freezing the backbone~\cite{vpt, vpt2}. As shown in~\cref{fig:prompt}, this strategy greatly reduces trainable parameters and improves efficiency~\cite{E2vpt}. It has been shown to enhance model interpretability and fine-grained recognition via class-specific prompts~\cite{promptcam}. However, PT still assumes white-box access to model parameters and requires supervision, making it unsuitable in label-scarce or black-box environments.

\subsection{Prompt Tuning in Black-box Setting}
To overcome gradient restrictions, recent methods introduce \textit{gradient-free prompt tuning} for black-box models. As illustrated in~\cref{fig:blackbox}, BlackVIP~\cite{blackvip} and BAPs~\cite{BAPs} optimize prompts directly in the input space to manipulate downstream outputs for classification and segmentation via zeroth-order optimization \cite{blackvip}. CraFT~\cite{CraFT} extends this by combining input prompts (optimized via CMA-ES) and a refinement module (trained via gradients on logits).

To reduce reliance on labels, VDPG~\cite{vdpg} and L2C~\cite{l2c} propose learning \textit{domain prompt generators}, trained with gradients from a refinement module, to adapt black-box features without ground-truth supervision. However, these methods assume: (1) access to features or logits; (2) pretrained \textcolor{red}{robust} black-box downstream models (e.g., CLIP \cite{clip}); and (3) in the case of VDPG and L2C, multiple source domains for domain-generalizable prompt generation. These assumptions are impractical in real-world, privacy-constrained environments such as healthcare.

\subsection{Prompt Tuning in Inference-blackbox Setting}
The inference-blackbox setting, illustrated in~\cref{fig:inference}, is the most restrictive scenario, where only the final predictions, \textbf{including image class IDs and segmentation masks (optional)}, are provided from the black-box downstream model. \textbf{NO} gradients, intermediate features, logits, and model parameters are accessible—conditions often encountered in real-world healthcare deployments.

To address this challenge, we propose \textbf{UI-Styler}, a prompt tuning framework designed explicitly for the inference-blackbox regime. Unlike previous approaches that still require supervision or logits~\cite{blackvip, CraFT}, UI-Styler leverages unpaired target samples and pseudo labels to drive adaptation via class-aware prompts. Our method operates entirely in the input space and applies a dual-level stylization strategy, aligning source images with the target domain in both appearance and semantics.



\section{Detailed Content and Style Losses} \label{sec:loss}

Following style transfer works~\cite{StyTr2, s2wat, SANet}, we adopt perceptual losses computed from a pre-trained VGG-19 network to guide structural preservation and appearance alignment.

\noindent \textbf{Content Loss.}
The content loss $\mathcal{L}_c$ measures the $\ell_2$ distance between the feature representations of the stylized image $\widetilde{x}_s$ and the original source image $x_s$, extracted from two higher-level layers of VGG-19:
\begin{equation}
\mathcal{L}_c = \left\| \phi^{4,1}(\widetilde{x}_s) - \phi^{4,1}(x_s) \right\|_2^2 
+ \left\| \phi^{5,1}(\widetilde{x}_s) - \phi^{5,1}(x_s) \right\|_2^2,
\end{equation}
where $\phi^{l,1}(\cdot)$ denotes the activation from the first convolutional layer after the $l$-th ReLU block.

\noindent \textbf{Style Loss.}
To capture multi-scale stylistic characteristics, we define the style loss $\mathcal{L}_s$ using the mean and standard deviation statistics of VGG features from multiple layers:
\begin{equation}
\begin{aligned}
\mathcal{L}_s = \sum_{l=2}^{5} \big(
& \left\| \mu(\phi^{l,1}(\widetilde{x}_s)) - \mu(\phi^{l,1}(x_t)) \right\|_2^2 \\
+ & \left\| \sigma(\phi^{l,1}(\widetilde{x}_s)) - \sigma(\phi^{l,1}(x_t)) \right\|_2^2
\big),
\end{aligned}
\end{equation}
\noindent where $\mu(\cdot)$ and $\sigma(\cdot)$ represent the mean and standard deviation of the extracted features, respectively.

\begin{table}[!b]
\centering
\resizebox{.9\columnwidth}{!}{%
\begin{tabular}{c||cccc}
\toprule
Target Domains & Acc\textcolor{red}{$\uparrow$} & AUC\textcolor{red}{$\uparrow$} & Dice\textcolor{red}{$\uparrow$} & IoU\textcolor{red}{$\uparrow$} \\

\midrule

BUSBRA \cite{BUS-BRA} & 89.17 & 94.71 & 90.99 & 84.16 \\
BUSI \cite{BUSI} & 92.82 & 96.09 & 86.63 & 78.53 \\
UCLM \cite{UCLM} & 93.75 & 97.63 & 88.28 & 80.31 \\
UDIAT \cite{UDIAT} & 91.84 & 97.65 & 90.51 & 83.29 \\

\bottomrule
\end{tabular}%
}
\caption{%
\textbf{Downstream Performance on Target Domains.}
We report the performance of the black-box downstream models on each domain for reference. The results are evaluated on the $30\%$ \textbf{testing} set. 
The high classification/segmentation performance indicates that these black-box downstream models are reliable enough to deploy clinical diagnosis applications.
}
\label{tab:target_perf}
\end{table}

\section{Downstream Performance on Target Domains} \label{sec:target_perf}

To provide reference results, we report the performance of the black-box downstream model when directly evaluated on each target domain with the $30\%$ \textbf{testing} set.

As listed in \cref{tab:target_perf}, the black-box model delivers strong performance on all target domains, with accuracy above $89\%$ and AUC consistently exceeding $94\%$. Segmentation results are also reliable, as Dice scores remain above $86\%$ and IoU above $78\%$ across all cases. 
These results confirm that the black-box downstream model can serve to evaluate unpaired image translation methods in cross-domain tasks. Furthermore, its reliable performance suggests suitability for deploying clinical diagnosis applications.

\section{Additional Experiments} \label{sec:extra_exp}

\begin{table*}[!t]
\centering
\resizebox{\textwidth}{!}{%
\begin{tabular}{
  c c||@{\hspace{.1em}}c@{\hspace{.1em}}|M M M M M|
  @{\hspace{.1em}}c@{\hspace{.1em}}|M M M M M|
  @{\hspace{.1em}}c@{\hspace{.1em}}|M M M M M@{\hspace{.1em}}}
\toprule
$\mathcal{L}_{\text{dir}}$ & $\mathcal{L}_{\text{sup}}$ & 
\textbf{Tasks} & 
KID\textcolor{red}{$\downarrow$} & Acc\textcolor{red}{$\uparrow$} & AUC\textcolor{red}{$\uparrow$} & Dice\textcolor{red}{$\uparrow$} & IoU\textcolor{red}{$\uparrow$} &
\textbf{Tasks} & 
KID\textcolor{red}{$\downarrow$} & Acc\textcolor{red}{$\uparrow$} & AUC\textcolor{red}{$\uparrow$} & Dice\textcolor{red}{$\uparrow$} & IoU\textcolor{red}{$\uparrow$} &
\textbf{Tasks} & 
KID\textcolor{red}{$\downarrow$} & Acc\textcolor{red}{$\uparrow$} & AUC\textcolor{red}{$\uparrow$} & Dice\textcolor{red}{$\uparrow$} & IoU\textcolor{red}{$\uparrow$} \\
\midrule

-- & \textcolor{red}{\checkmark} & \multirow{3.2}{*}{\shortstack{BUSBRA\\$\downarrow$\\BUSI}} & \underline{11.73} & 73.89 & 75.06 & 83.80 & 73.89 & \multirow{3.2}{*}{\shortstack{BUSBRA\\$\downarrow$\\UCLM}} & \underline{17.23} & \underline{74.96} & \underline{76.49} & 81.28 & 70.88 & \multirow{3.2}{*}{\shortstack{BUSBRA\\$\downarrow$\\UDIAT}} & \underline{12.11} & 65.90 & 68.83 & 85.82 & \underline{76.94} \\

\textcolor{red}{\checkmark} & -- & & 12.66 & \underline{75.13} & \underline{75.77} & \underline{84.47} & \textbf{74.80} & & 17.63 & 74.25 & 76.10 & \underline{81.74} & \underline{71.24} & & 12.71 & \underline{69.09} & \underline{70.84} & \underline{85.83} & 76.87 \\

\textcolor{red}{\checkmark} & \textcolor{red}{\checkmark} & & \textbf{11.20} & \textbf{75.84} & \textbf{76.33} & \textbf{84.52} & \underline{74.74} & & \textbf{16.91} & \textbf{75.13} & \textbf{76.78} & \textbf{82.06} & \textbf{71.73} & & \textbf{9.14} & \textbf{72.47} & \textbf{71.52} & \textbf{86.04} & \textbf{77.52} \\

\midrule

-- & \textcolor{red}{\checkmark} & \multirow{3.2}{*}{\shortstack{BUSI\\$\downarrow$\\BUSBRA}} & \textbf{10.50} & 84.10 & 87.12 & 83.01 & \underline{73.97} & \multirow{3.2}{*}{\shortstack{BUSI\\$\downarrow$\\UCLM}} & 12.43 & 70.77 & 74.91 & \underline{78.30} & 68.31 & \multirow{3.2}{*}{\shortstack{BUSI\\$\downarrow$\\UDIAT}} & 4.39 & \textbf{74.36} & 75.31 & \underline{80.30} & 71.21 \\

\textcolor{red}{\checkmark} & -- & & 12.74 & \underline{84.62} & \underline{87.22} & \underline{83.04} & 
\underline{73.97} & & \underline{11.25} & \underline{71.79} & \underline{76.13} & 78.13 & \underline{68.40} & & \underline{3.78} & \underline{73.85} & \underline{77.74} & 80.19 & \underline{71.27} \\

\textcolor{red}{\checkmark} & \textcolor{red}{\checkmark} & & \underline{11.25} & \textbf{85.13} & \textbf{88.14} & \textbf{83.15} & \textbf{74.05} & & \textbf{11.05} & \textbf{74.36} & \textbf{77.15} & \textbf{78.83} & \textbf{68.61} & & \textbf{3.61} & \textbf{74.36} & \textbf{78.89} & \textbf{80.49} & \textbf{71.61} \\

\midrule

-- & \textcolor{red}{\checkmark} & \multirow{3.2}{*}{\shortstack{UCLM\\$\downarrow$\\BUSBRA}} & \underline{10.22} & \underline{87.50} & \underline{92.49} & 81.71 & 71.60 & \multirow{3.2}{*}{\shortstack{UCLM\\$\downarrow$\\BUSI}} & \underline{13.13} & \underline{78.75} & \underline{83.43} & 78.82 & 68.69 & \multirow{3.2}{*}{\shortstack{UCLM\\$\downarrow$\\UDIAT}} & 15.76 & 62.50 & 65.58 & 82.97 & 72.91 \\

\textcolor{red}{\checkmark} & -- & & 12.91 & 83.75 & 91.01 & \underline{82.07} & \underline{71.71} & & 13.85 & 76.25 & 81.95 & \underline{79.67} & \underline{69.40} & & \underline{14.91} & \underline{65.00} & \underline{70.18} & \underline{83.02} & \underline{73.19} \\

\textcolor{red}{\checkmark} & \textcolor{red}{\checkmark} & & \textbf{9.60} & \textbf{88.75} & \textbf{94.93} & \textbf{82.79} & \textbf{72.65} & & \textbf{12.40} & \textbf{80.00} & \textbf{85.60} & \textbf{80.22} & \textbf{69.78} & & \textbf{13.56} & \textbf{71.25} & \textbf{73.36} & \textbf{83.16} & \textbf{73.27} \\

\midrule

-- & \textcolor{red}{\checkmark} & \multirow{3.2}{*}{\shortstack{UDIAT\\$\downarrow$\\BUSBRA}} & \underline{5.70} & \underline{83.67} & 76.07 & 88.32 & 79.99 & \multirow{3.2}{*}{\shortstack{UDIAT\\$\downarrow$\\BUSI}} & 4.73 & \underline{89.80} & \underline{93.80} & 83.36 & 73.89 & \multirow{3.2}{*}{\shortstack{UDIAT\\$\downarrow$\\UCLM}} & 16.02 & \underline{83.67} & \underline{85.47} & \underline{85.72} & \underline{76.21} \\

\textcolor{red}{\checkmark} & -- & & 6.71 & 81.63 & \underline{77.35} & \underline{88.38} & \underline{80.12} & & \underline{4.59} & \underline{89.80} & 92.95 & \underline{83.92} & \underline{74.52} & & \underline{13.03} & 81.63 & 81.84 & 85.32 & 75.87 \\

\textcolor{red}{\checkmark} & \textcolor{red}{\checkmark} & & \textbf{5.25} & \textbf{87.76} & \textbf{79.27} & \textbf{88.45} & \textbf{80.13} & & \textbf{4.47} & \textbf{91.84} & \textbf{96.15} & \textbf{85.39} & \textbf{76.09} & & \textbf{12.33} & \textbf{85.71} & \textbf{88.25} & \textbf{85.83} & \textbf{76.46} \\

\bottomrule
\end{tabular}
}
\vspace{-.5em}
\caption{\textbf{Ablation Study on Loss Contributions.} We evaluate the impact of $\mathcal{L}_{\text{dir}}$ and $\mathcal{L}_{\text{sup}}$ in the final objective across $12$ cross-device ultrasound tasks. Each result is reported under $5$ metrics: KID, Acc, AUC, Dice, and IoU. \textbf{Bold} denotes the best result, and \underline{underline} indicates the second-best.}
\label{tab:loss_ablation}
\end{table*}

\subsection{Ablation Study on Loss Contributions} \label{subsec:contri_loss}

Since the content loss ($\mathcal{L}_c$) and style loss ($\mathcal{L}_s$) are standard components in style transfer frameworks, we focus on evaluating the additional contributions of the proposed direction loss ($\mathcal{L}_{\text{dir}}$) and supervised loss ($\mathcal{L}_{\text{sup}}$), as reported in \cref{tab:loss_ablation}. Specifically, we find that using only $\mathcal{L}_{\text{sup}}$—\textit{without the explicit guidance from $\mathcal{L}_{\text{dir}}$}—often causes the stylized features ($\widetilde{\mathcal{F}}_{s \rightarrow t}$) to be matched with \textit{incorrect} class-specific prompts ($\mathcal{P}_c$). From \cref{tab:loss_ablation}, we observe that the accuracy drops \textbf{drastically} from $71.25$ (full setting) to $62.50$ in the UCLM$\rightarrow$UDIAT task.

Moreover, when using only $\mathcal{L}_{\text{dir}}$—\textit{without the supervision from} $\mathcal{L}_{\text{sup}}$—the prompts lack supervision from the target domain and thus fail to learn class-specific characteristics. As a result, in the UDIAT$\rightarrow$BUSI task, the Dice score declines from $85.39$ to $83.92$, and the AUC drops from $96.15$ to $92.95$.

Consequently, the superior performance achieved with the full setting of $\mathcal{L}_{\text{dir}}$ and $\mathcal{L}_{\text{sup}}$ provides strong evidence that the stylized features ($\widetilde{\mathcal{F}}_{s \rightarrow t}$) are effectively aligned with the correct class while preserving diagnostic traits.

\begin{table}[!t]
\centering
\resizebox{\columnwidth}{!}{%
\begin{tabular}{@{\hspace{.1em}}c@{\hspace{.1em}}cc|c@{\hspace{.8em}}c||ccccc}
\toprule

& $\lambda_{c}$ & $\lambda_{s}$ & $\lambda_{\text{dir}}$ & $\lambda_{\text{sup}}$ & KID\textcolor{red}{$\downarrow$} & Acc\textcolor{red}{$\uparrow$} & AUC\textcolor{red}{$\uparrow$} & Dice\textcolor{red}{$\uparrow$} & IoU\textcolor{red}{$\uparrow$} \\

\midrule


\multirow{2}{*}{\rotatebox{90}{G\textbf{(1)}}} & \textcolor{red}{2} & 1 & 1 & 1 & 12.38 & 77.71 & 80.53 & 82.90 & 73.36  \\

& 1 & \textcolor{red}{2} & 1 & 1 & \textbf{8.75} & 78.20 & \underline{80.75} & 82.77 & 73.12  \\

\midrule

\multirow{2}{*}{\rotatebox{90}{G\textbf{(2)}}} & 1 & 1 & \textcolor{red}{2} & 1 & 10.62 & \underline{79.71} & 80.20 & 82.96 & \underline{73.44}  \\

& 1 & 1 & 1 & \textcolor{red}{2} & 10.40 & 78.12 & 80.65 & \underline{82.97} & \underline{73.44}  \\

\midrule

\rowcolor{green!10} & 1 & 1 & 1 & 1 & \underline{10.06} & \textbf{80.22} & \textbf{82.20} & \textbf{83.41} & \textbf{73.89} \\
\bottomrule
\end{tabular}%
}
\vspace{-.5em}
\caption{%
\textbf{Loss Weight Configurations.} 
We report the \textit{averaged results} of different loss weight configurations over $12$ cross-device tasks under $5$ metrics: KID, Acc, AUC, Dice, and IoU. 
\textbf{Bold} denotes the best result, and \underline{underline} indicates the second-best. 
\textit{The per-task results are reported in \cref{tab:loss_weight}}.
}
\label{tab:loss_average}
\end{table}

\subsection{Loss Weight Configurations} \label{subsec:weight_loss}

We investigate different combinations of loss functions across $12$ cross-device tasks. Since the content loss ($\mathcal{L}_c$) and style loss ($\mathcal{L}_s$) are the baseline objectives in the style transfer process, we divide the study into \textbf{two main groups} (G)  with distinct optimization goals: 
\textbf{(1)} \textit{style transfer}, where $\mathcal{L}_c$ and $\mathcal{L}_s$ are computed to guide the transformation $\big(I^{s-style}_{\textcolor{red}{s-content}}, I^{\textcolor{teal}{t-style}}_{t-content}\big) \rightarrow I^{\textcolor{teal}{t-style}}_{\textcolor{red}{s-content}}$; and
\textbf{(2)} \textit{prompt learning}, where the direction loss $\mathcal{L}_{\text{dir}}$ and the supervised loss $\mathcal{L}_{\text{sup}}$ are used to optimize the template prompt set $\mathcal{P}$. For each group, we assess three pairwise settings—$(1,1)$, $(2,1)$, and $(1,2)$—with the averaged results in \cref{tab:loss_average}.

\underline{For the G\textbf{(1)}}, we find that increasing $\mathcal{L}_c$ tends to overshadow $\mathcal{L}_s$, resulting in insufficient transfer of the target style, especially when the domain gap is large. Conversely, increasing $\mathcal{L}_s$ may over-stylize the content information, causing content degradation. Therefore, balancing content and style information proves essential, yielding improvements across all metrics.
\underline{In the G\textbf{(2)}}, we observe that balancing $\mathcal{L}_{\text{dir}}$ and $\mathcal{L}_{\text{sup}}$ yields consistently higher Acc, AUC, Dice, and IoU compared to overwhelming-weight settings. This trend can be further explained by examining the effect of unbalanced weights: when $\mathcal{L}_{\text{dir}}$  dominates, prompt learning leans toward directional alignment but lacks pseudo target label guidance, reducing discriminability. Conversely, increasing $\mathcal{L}_{\text{sup}}$, the supervision from pseudo target labels overshadows the correlation-alignment effect of $\mathcal{L}_{\text{dir}}$, thereby limiting the selection of suitable class-specific prompts, $\mathcal{P}_c$.

Based on these findings, the balanced loss weighting provides the most reliable performance, achieving $4/5$ best metrics, including Acc of $80.22$, AUC of $82.20$, Dice of $83.41$, and IoU of $73.89$. \textit{For a comprehensive comparison, we provide the per-task results in \cref{tab:loss_weight}}.

\begin{table*}[!t]
\centering
\resizebox{\textwidth}{!}{%
\begin{tabular}{
@{\hspace{.2em}}c@{\hspace{.2em}}c@{\hspace{.1em}}|@{\hspace{.1em}}c@{\hspace{.2em}}c@{\hspace{.2em}}||@{\hspace{.1em}}c@{\hspace{.1em}}|M M M M M|
@{\hspace{.1em}}c@{\hspace{.1em}}|M M M M M|
@{\hspace{.1em}}c@{\hspace{.1em}}|M M M M M@{\hspace{.1em}}}
\toprule

$\lambda_{c}$ & $\lambda_{s}$ & $\lambda_{\text{dir}}$ & $\lambda_{\text{sup}}$ & 
\textbf{Tasks} & 
KID\textcolor{red}{$\downarrow$} & Acc\textcolor{red}{$\uparrow$} & AUC\textcolor{red}{$\uparrow$} & Dice\textcolor{red}{$\uparrow$} & IoU\textcolor{red}{$\uparrow$} &
\textbf{Tasks} & 
KID\textcolor{red}{$\downarrow$} & Acc\textcolor{red}{$\uparrow$} & AUC\textcolor{red}{$\uparrow$} & Dice\textcolor{red}{$\uparrow$} & IoU\textcolor{red}{$\uparrow$} &
\textbf{Tasks} & 
KID\textcolor{red}{$\downarrow$} & Acc\textcolor{red}{$\uparrow$} & AUC\textcolor{red}{$\uparrow$} & Dice\textcolor{red}{$\uparrow$} & IoU\textcolor{red}{$\uparrow$} \\
\midrule

\textcolor{red}{2} & 1 & 1 & 1 & \multirow{5.2}{*}{\shortstack{BUSBRA\\$\downarrow$\\BUSI}} & 15.45 & 75.31 & 74.67 & 84.41 & 74.69 & \multirow{5.2}{*}{\shortstack{BUSBRA\\$\downarrow$\\UCLM}} & 16.30 & \textbf{75.13} & 75.76 & 82.15 & 71.74 & \multirow{5.2}{*}{\shortstack{BUSBRA\\$\downarrow$\\UDIAT}} & 13.90 & 68.21 & 70.73 & \underline{85.97} & \underline{77.03} \\

1 & \textcolor{red}{2} & 1 & 1 & & \textbf{8.46} & 73.00 & \underline{75.69} & 83.76 & 73.89 & & \textbf{13.39} & 74.60 & 76.61 & 82.16 & 71.84 & & \textbf{8.87} & 67.14 & 68.68 & 85.87 & 76.93 \\

1 & 1 & \textcolor{red}{2} & 1 & & 13.06 & \underline{75.49} & 75.55 & \underline{84.43} & \underline{74.71} & & \underline{15.05} & \underline{74.96} & \textbf{76.98} & \underline{82.29} & \underline{71.90} & & 12.48 & \underline{69.09} & \underline{71.13} & 85.83 & 76.85 \\

1 & 1 & 1 & \textcolor{red}{2} & & 13.25 & 74.25 & 74.28 & 84.25 & 74.46 & & 16.71 & 74.42 & 76.30 & \textbf{82.41} & \textbf{72.09} & & 12.52 & 67.50 & 70.67 & 85.80 & 76.85 \\

1 & 1 & 1 & 1 & & \underline{11.20} & \textbf{75.84} & \textbf{76.33} & \textbf{84.52} & \textbf{74.74} & & 16.91 & \textbf{75.13} & \underline{76.78} & 82.06 & 71.73 & & \underline{9.14} & \textbf{72.47} & \textbf{71.52} & \textbf{86.04} & \textbf{77.52} \\

\midrule

\textcolor{red}{2} & 1 & 1 & 1 & \multirow{5.2}{*}{\shortstack{BUSI\\$\downarrow$\\BUSBRA}} & 10.89 & \underline{84.62} & \underline{88.09} & \textbf{83.27} & \textbf{74.17} & \multirow{5.2}{*}{\shortstack{BUSI\\$\downarrow$\\UCLM}} & 12.52 & 70.77 & 75.67 & 77.93 & 68.20 & \multirow{5.2}{*}{\shortstack{BUSI\\$\downarrow$\\UDIAT}} & 4.29 & 73.85 & 76.96 & \underline{80.40} & \underline{71.41} \\

1 & \textcolor{red}{2} & 1 & 1 & & \textbf{5.52} & 82.56 & 86.22 & 83.08 & 73.84 & & \textbf{10.84} & \textbf{75.90} & \textbf{78.07} & \underline{78.34} & \textbf{68.62} & & \textbf{3.43} & \underline{74.36} & 76.05 & 79.77 & 70.68 \\

1 & 1 & \textcolor{red}{2} & 1 & & 7.61 & \textbf{85.13} & 87.16 & 82.86 & 73.89 & & \underline{10.92} & 72.82 & 75.87 & 77.96 & 68.29 & & 4.45 & \textbf{75.38} & 76.49 & 80.35 & 71.40 \\

1 & 1 & 1 & \textcolor{red}{2} & & \underline{7.46} & \textbf{85.13} & 88.05 & 82.74 & 73.57 & & 11.98 & \underline{74.36} & 75.91 & 78.12 & 68.55 & & 3.69 & 73.85 & \underline{78.46} & 80.19 & 71.18 \\

1 & 1 & 1 & 1 & & 11.25 & \textbf{85.13} & \textbf{88.14} & \underline{83.15} & \underline{74.05} & & 11.05 & \underline{74.36} & \underline{77.15} & \textbf{78.83} & \underline{68.61} & & \underline{3.61} & \underline{74.36} & \textbf{78.89} & \textbf{80.49} & \textbf{71.61} \\

\midrule

\textcolor{red}{2} & 1 & 1 & 1 & \multirow{5.2}{*}{\shortstack{UCLM\\$\downarrow$\\BUSBRA}} & 15.02 & 86.25 & \underline{93.37} & 81.67 & 71.70 & \multirow{5.2}{*}{\shortstack{UCLM\\$\downarrow$\\BUSI}} & 14.85 & 75.00 & 83.77 & 78.63 & 68.31 & \multirow{5.2}{*}{\shortstack{UCLM\\$\downarrow$\\UDIAT}} & 16.17 & 66.25 & 72.62 & \underline{82.80} & \underline{72.82} \\

1 & \textcolor{red}{2} & 1 & 1 & & \textbf{8.98} & 85.00 & 93.31 & 81.73 & 71.65 & & \textbf{11.84} & \textbf{80.00} & \underline{84.18} & 78.40 & 67.95 & & 15.09 & \underline{68.75} & 70.39 & 82.76 & 72.64 \\

1 & 1 & \textcolor{red}{2} & 1 & & 11.82 & \textbf{90.00} & 93.31 & 82.73 & 72.32 & & 13.57 & \underline{77.50} & 82.76 & \underline{79.32} & \underline{69.20} & & 14.20 & \underline{68.75} & 71.26 & 82.62 & 72.75 \\

1 & 1 & 1 & \textcolor{red}{2} & & 12.02 & 85.00 & 91.55 & \textbf{83.01} & \textbf{72.75} & & \underline{12.27} & 76.25 & 82.35 & 78.82 & 68.59 & & \textbf{13.13} & 67.50 & \textbf{73.83} & 82.76 & 72.77 \\

1 & 1 & 1 & 1 & & \underline{9.60} & \underline{88.75} & \textbf{94.93} & \underline{82.79} & \underline{72.65} & & 12.40 & \textbf{80.00} & \textbf{85.60} & \textbf{80.22} & \textbf{69.78} & & \underline{13.56} & \textbf{71.25} & \underline{73.36} & \textbf{83.16} & \textbf{73.27} \\

\midrule

\textcolor{red}{2} & 1 & 1 & 1 & \multirow{5.2}{*}{\shortstack{UDIAT\\$\downarrow$\\BUSBRA}} & 7.07 & 83.67 & \textbf{79.49} & 88.19 & \underline{79.84} & \multirow{5.2}{*}{\shortstack{UDIAT\\$\downarrow$\\BUSI}} & 4.27 & 89.80 & 92.52 & 83.95 & 74.37 & \multirow{5.2}{*}{\shortstack{UDIAT\\$\downarrow$\\UCLM}} & 17.78 & 83.67 & 82.69 & 85.42 & 75.98 \\

1 & \textcolor{red}{2} & 1 & 1 & & \textbf{3.13} & 83.67 & 77.78 & 88.04 & 79.56 & & \textbf{3.13} & 89.80 & \underline{94.02} & \underline{84.14} & 74.26 & & \underline{12.32} & 83.67 & \underline{88.03} & 85.14 & 75.63 \\

1 & 1 & \textcolor{red}{2} & 1 & & 5.62 & \underline{85.71} & 76.71 & 87.94 & 79.63 & & \underline{4.16} & \textbf{93.88} & 90.81 & 83.57 & 74.15 & & 14.55 & \textbf{87.76} & 84.40 & \underline{85.57} & \underline{76.15} \\

1 & 1 & 1 & \textcolor{red}{2} & & 5.93 & 83.67 & 78.21 & \underline{88.37} & \textbf{80.13} & & 4.35 & \underline{91.84} & 92.95 & 83.89 & \underline{74.49} & & \textbf{11.47} & 83.67 & 85.26 & 85.31 & 75.86 \\

1 & 1 & 1 & 1 & & \underline{5.25} & \textbf{87.76} & \underline{79.27} & \textbf{88.45} & \textbf{80.13} & & 4.47 & \underline{91.84} & \textbf{96.15} & \textbf{85.39} & \textbf{76.09} & & 12.33 & \underline{85.71} & \textbf{88.25} & \textbf{85.83} & \textbf{76.46} \\

\bottomrule
\end{tabular}
}
\vspace{-.5em}
\caption{%
\textbf{Loss Weight Configurations.} 
We report the per-task performance of different loss weight configurations across $12$ cross-device tasks, evaluated under $5$ metrics: KID, Acc, AUC, Dice, and IoU. 
\textbf{Bold} denotes the best result, and \underline{underline} indicates the second-best.
}
\label{tab:loss_weight}
\end{table*}

\subsection{Sensitivity of Pattern-matching Parameters} \label{subsec:PM_params}

We analyze the sensitivity of our pattern-matching module with respect to the number of ViT blocks as shown in \cref{tab:avg_blocks}, which reports the averaged results over $12$ cross-device tasks. The floating-point operations (FLOPs) are measured with an input image size of $256 \times 256$. We observe that the configuration with $3$ ViT blocks achieves the best overall trade-off, obtaining the lowest KID ($10.06$) and highest Acc ($80.22$). Specifically, compared to $5$ blocks, the performance gap is marginal (only $0.37$ in AUC and $0.16$ in Dice), while the FLOPs are reduced from $64.30$G to $55.70$G. More importantly, compared to the $2$-block setting, $3$ blocks show a substantial improvement of $2.48\%$ in Acc (from $77.74$ to $80.22$) and consistent gains across other metrics.  

These results indicate that using $3$ ViT blocks provides the most efficient balance between computational cost and performance. Hence, we adopt $3$ blocks as the default configuration of the pattern-matching module. \textit{For comprehensive comparison, we also provide the per-task performance in \cref{tab:blocks}.}

\begin{table}[!t]
\centering
\resizebox{\columnwidth}{!}{%
\begin{tabular}{c||cccccc}
\toprule
$\textcolor{red}{\#}$Blocks & KID\textcolor{red}{$\downarrow$} & Acc\textcolor{red}{$\uparrow$} & AUC\textcolor{red}{$\uparrow$} & Dice\textcolor{red}{$\uparrow$} & IoU\textcolor{red}{$\uparrow$} & FLOPs\textcolor{red}{$\downarrow$} \\

\midrule

2 & \underline{10.07} & 77.74 & 79.89 & 82.85 & 73.30 & \textbf{51.40G} \\
\rowcolor{green!10} 3 & \textbf{10.06} & \textbf{80.22} & \underline{82.20} & \underline{83.41} & \underline{73.89} & \underline{55.70G} \\
5 & 10.61 & \underline{80.21} & \textbf{82.57} & \textbf{83.57} & \textbf{73.97} & 64.30G \\

\bottomrule
\end{tabular}%
}
\vspace{-.5em}
\caption{%
\textbf{Sensitivity of Pattern-matching Parameters.} 
We present the \textit{average performance} of different numbers of ViT blocks in the pattern-matching module across $12$ cross-device tasks, evaluated on $5$ metrics (KID, Acc, AUC, Dice, IoU) and computational cost (FLOPs). 
\textbf{Bold} denotes the best result, and \underline{underline} indicates the second-best. 
\textit{The per-task results are reported in \cref{tab:blocks}}.
}
\vspace{-.5em}
\label{tab:avg_blocks}
\end{table}

\begin{table*}[!t]
\centering
\resizebox{\textwidth}{!}{%
\begin{tabular}{
@{\hspace{.1em}}c@{\hspace{.1em}}||@{\hspace{.1em}}c@{\hspace{.1em}}|M M M M M|
  @{\hspace{.1em}}c@{\hspace{.1em}}|M M M M M|
  @{\hspace{.1em}}c@{\hspace{.1em}}|M M M M M@{\hspace{.1em}}}
\toprule
$\textcolor{red}{\#}$Blocks & 
\textbf{Tasks} & 
KID\textcolor{red}{$\downarrow$} & Acc\textcolor{red}{$\uparrow$} & AUC\textcolor{red}{$\uparrow$} & Dice\textcolor{red}{$\uparrow$} & IoU\textcolor{red}{$\uparrow$} &
\textbf{Tasks} & 
KID\textcolor{red}{$\downarrow$} & Acc\textcolor{red}{$\uparrow$} & AUC\textcolor{red}{$\uparrow$} & Dice\textcolor{red}{$\uparrow$} & IoU\textcolor{red}{$\uparrow$} &
\textbf{Tasks} & 
KID\textcolor{red}{$\downarrow$} & Acc\textcolor{red}{$\uparrow$} & AUC\textcolor{red}{$\uparrow$} & Dice\textcolor{red}{$\uparrow$} & IoU\textcolor{red}{$\uparrow$} \\
\midrule

2 & \multirow{3.2}{*}{\shortstack{BUSBRA\\$\downarrow$\\BUSI}} & 12.17 & 74.78 & 74.33 & 83.92 & 74.05 & \multirow{3.2}{*}{\shortstack{BUSBRA\\$\downarrow$\\UCLM}} & \textbf{14.40} & 74.07 & \underline{77.27} & \underline{82.08} & 71.68 & \multirow{3.2}{*}{\shortstack{BUSBRA\\$\downarrow$\\UDIAT}} & \underline{11.67} & 66.61 & 68.56 & 85.78 & 76.83 \\

3 & & \textbf{11.20} & \underline{75.84} & \underline{76.33} & \textbf{84.52} & \textbf{74.74} & & 16.91 & \underline{75.13} & 76.78 & 82.06 & \underline{71.73} & & \textbf{9.14} & \textbf{72.47} & \underline{71.52} & \underline{86.04} & \underline{77.52} \\

5 & & \underline{11.87} & \textbf{76.55} & \textbf{77.40} & \underline{84.24} & \underline{74.46} & & \underline{15.19} & \textbf{77.62} & \textbf{78.30} & \textbf{82.29} & \textbf{71.98} & & 13.61 & \underline{69.45} & \textbf{72.52} & \textbf{86.83} & \textbf{77.80} \\

\midrule

2 & \multirow{3.2}{*}{\shortstack{BUSI\\$\downarrow$\\BUSBRA}} & \underline{7.04} & 83.59 & 86.33 & 83.14 & 74.03 & \multirow{3.2}{*}{\shortstack{BUSI\\$\downarrow$\\UCLM}} & \textbf{10.67} & \underline{73.85} & 75.76 & 77.80 & 68.08 & \multirow{3.2}{*}{\shortstack{BUSI\\$\downarrow$\\UDIAT}} & \underline{4.12} & \underline{74.36} & 77.63 & \underline{80.21} & 71.09 \\

3 & & 11.25 & \textbf{85.13} & \underline{88.14} & \underline{83.15} & \underline{74.05} & & 11.05 & \textbf{74.36} & \underline{77.15} & \textbf{78.83} & \textbf{68.61} & & \textbf{3.61} & \underline{74.36} & \textbf{78.89} & \textbf{80.49} & \underline{71.61} \\

5 & & \textbf{6.43} & \underline{84.62} & \textbf{89.17} & \textbf{83.20} & \textbf{74.26} & & \underline{11.02} & \textbf{74.36} & \textbf{79.20} & \underline{78.07} & \underline{68.35} & & 4.29 & \textbf{76.41} & \underline{78.62} & \textbf{80.49} & \textbf{72.48} \\

\midrule

2 & \multirow{3.2}{*}{\shortstack{UCLM\\$\downarrow$\\BUSBRA}} & \underline{12.24} & \underline{86.25} & 93.44 & 82.54 & 72.64 & \multirow{3.2}{*}{\shortstack{UCLM\\$\downarrow$\\BUSI}} & 12.82 & \underline{77.50} & 82.08 & 78.35 & 67.91 & \multirow{3.2}{*}{\shortstack{UCLM\\$\downarrow$\\UDIAT}} & \textbf{12.94} & 68.75 & 71.33 & 82.61 & 72.65 \\

3 & & \textbf{9.60} & \textbf{88.75} & \textbf{94.93} & \underline{82.79} & \underline{72.65} & & \textbf{12.40} & \textbf{80.00} & \underline{85.60} & \textbf{80.22} & \textbf{69.78} & & 13.56 & \underline{71.25} & \underline{73.36} & \underline{83.16} & \textbf{73.27} \\

5 & & 13.45 & \textbf{88.75} & \underline{94.46} & \textbf{83.05} & \textbf{72.86} & & \underline{12.57} & \textbf{80.00} & \textbf{85.73} & \underline{79.71} & \underline{69.28} & & \underline{13.20} & \textbf{71.50} & \textbf{74.92} & \textbf{83.96} & \underline{73.06} \\

\midrule

2 & \multirow{3.2}{*}{\shortstack{UDIAT\\$\downarrow$\\BUSBRA}} & 7.26 & 83.67 & \underline{77.99} & \underline{88.73} & \underline{80.58} & \multirow{3.2}{*}{\shortstack{UDIAT\\$\downarrow$\\BUSI}} & \textbf{3.78} & \underline{89.80} & 90.38 & 84.37 & 74.99 & \multirow{3.2}{*}{\shortstack{UDIAT\\$\downarrow$\\UCLM}} & \textbf{11.69} & \underline{79.59} & 83.55 & 84.64 & 75.08 \\

3 & & \textbf{5.25} & \textbf{87.76} & \textbf{79.27} & 88.45 & 80.13 & & 4.47 & \textbf{91.84} & \textbf{96.15} & \underline{85.39} & \textbf{76.09} & & \underline{12.33} & \textbf{85.71} & \underline{88.25} & \underline{85.83} & \underline{76.46} \\

5 & & \underline{6.56} & \underline{85.71} & \underline{77.99} & \textbf{88.96} & \textbf{80.78} & & \underline{4.21} & \textbf{91.84} & \underline{94.02} & \textbf{85.94} & \underline{75.51} & & 14.91 & \textbf{85.71} & \textbf{88.48} & \textbf{86.11} & \textbf{76.81} \\

\bottomrule
\end{tabular}
}
\caption{%
\textbf{Sensitivity of Pattern-matching Parameters.} 
We report the per-task performance of different numbers of ViT blocks in the pattern-matching module across $12$ cross-device ultrasound tasks, under $5$ metrics: KID, Acc, AUC, Dice, and IoU. 
\textbf{Bold} denotes the best result, and \underline{underline} indicates the second-best.
}
\label{tab:blocks}
\end{table*}

\section{Additional Analyses} \label{sec:extra_analysis}

\subsection{Comparison on Diagnostic Semantics} \label{subsec:diagnostic}

To demonstrate the capability of UI-Styler in preserving diagnostic semantics, we conduct a qualitative comparison of stylized results produced by unpaired image translation methods. Each comparison is performed on the same source image from BUSBRA with target-style counterparts from BUSI, UCLM, and UDIAT. According to the medical ultrasound literature \cite{ultrasound_diagnosis1, ultrasound_diagnosis2, ultrasound_diagnosis3}, the tumor region is a critical feature for accurate diagnosis.

\begin{figure*}[!t]
\centering
\includegraphics[width=\linewidth]{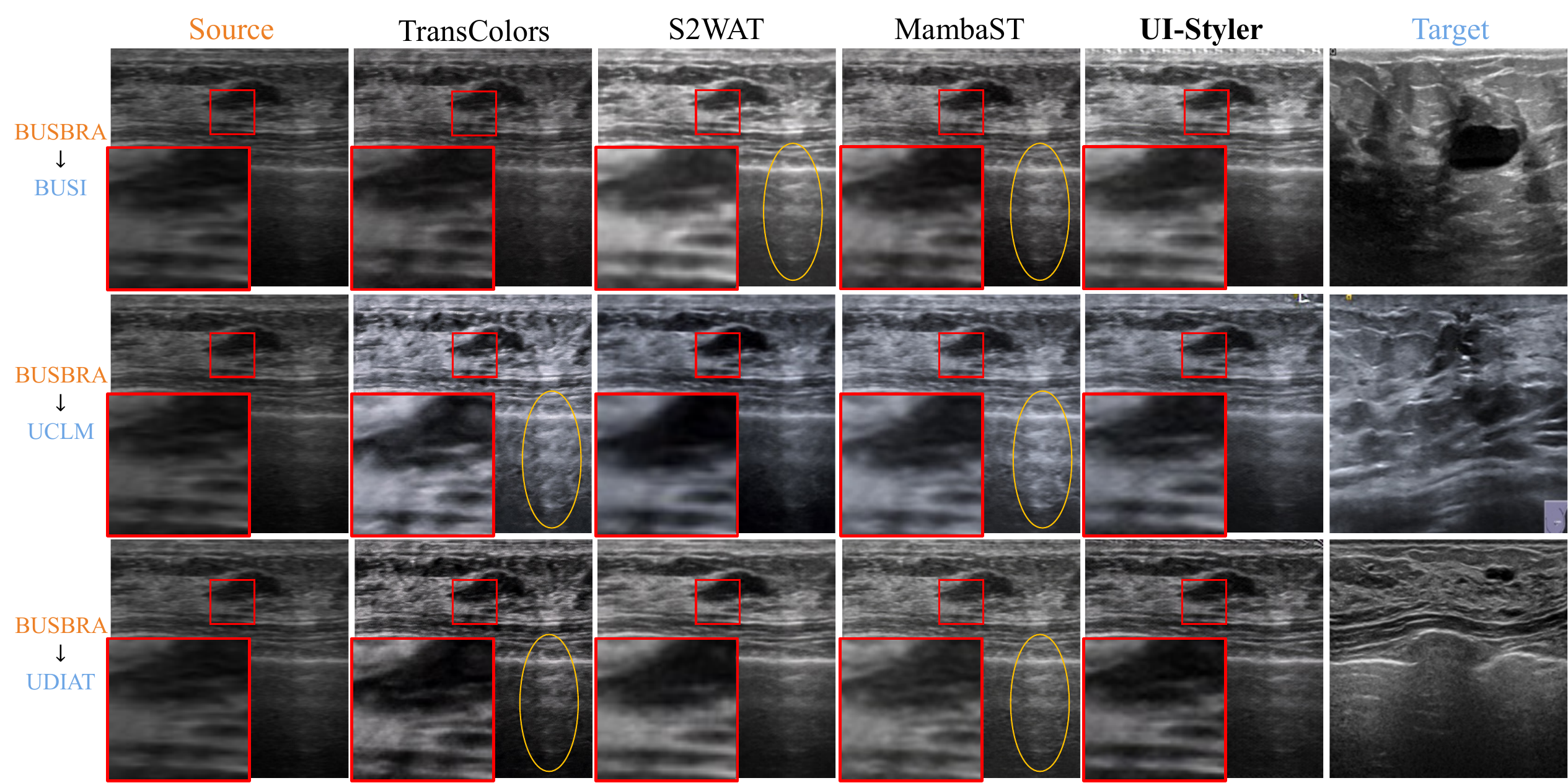}
\vspace{-1em}
\caption{%
\textbf{Comparison on Diagnostic Semantics.} 
We show stylized outputs from unpaired image translation methods, where each row displays the results generated from the same source-content image alongside target-style counterparts. Red boxes \textcolor{red}{$\square$} indicate zoomed tumor regions, while yellow ellipses \textcolor{myorange}{\scalebox{2}[1]{\ding{109}}} highlight artifact areas where competing methods fail to preserve diagnostic semantics. \textit{Please zoom in to view details more easily}.
} \label{fig:visual_evidence}
\end{figure*}

As shown in \cref{fig:visual_evidence}, previous methods often produce \textbf{inconsistencies} in tumor areas (highlighted by red boxes \textcolor{red}{$\square$}), as they mainly operate at the domain level, which imposes the target style onto the source content. As a result, different target devices can yield varying outcomes even for the same source image. In contrast, UI-Styler consistently preserves tumor regions across all tasks, providing strong evidence of its ability to maintain diagnostic semantics when incorporating class-aware transfer. 

Furthermore, competing approaches tend to generate undesired artifacts (marked by yellow ellipses \textcolor{myorange}{\scalebox{2}[1]{\ding{109}}}), whereas UI-Styler remains unaffected.

\begin{figure*}[!t]
\centering
\begin{subfigure}[b]{0.27\linewidth}
    \includegraphics[width=\linewidth]{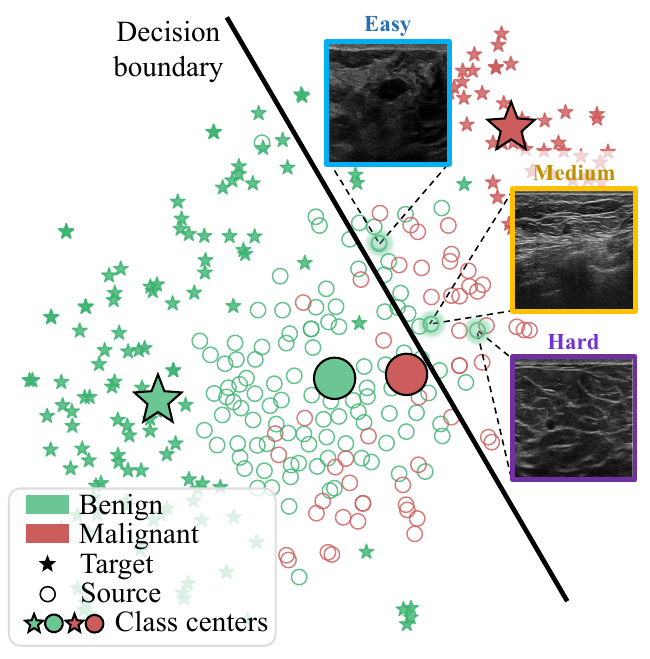}
    \subcaption{Before style transfer.}
    \label{fig:tsne_a}
\end{subfigure}
\hfill
\begin{subfigure}[b]{0.28\linewidth}
    \includegraphics[width=\linewidth]{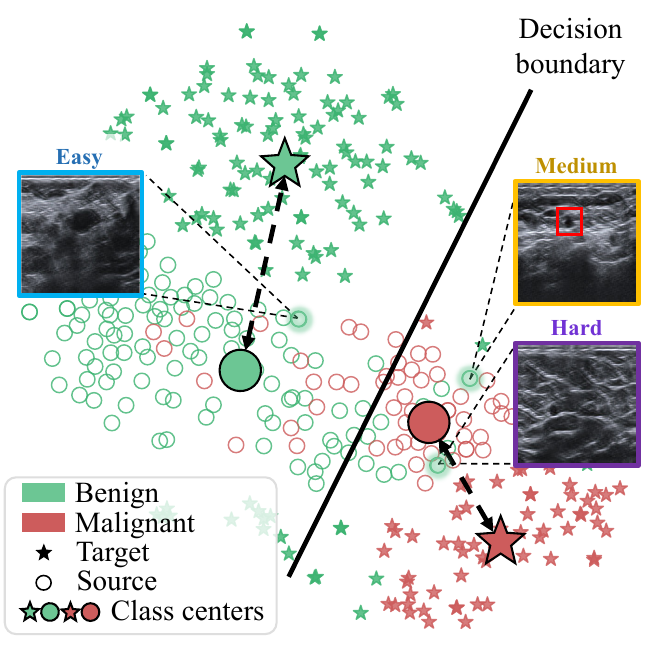}
    \subcaption{Only domain level.}
    \label{fig:tsne_b}
\end{subfigure}
\hfill
\begin{subfigure}[b]{0.27\linewidth}
    \includegraphics[width=\linewidth]{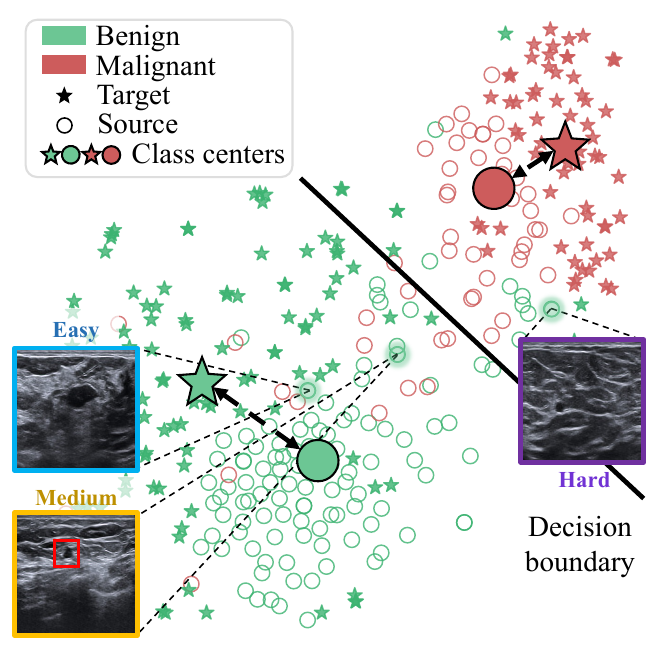}
    \subcaption{Domain and class levels.}
    \label{fig:tsne_c}
\end{subfigure}
\hfill
\begin{subfigure}[b]{0.13\linewidth}
    \includegraphics[width=\linewidth]{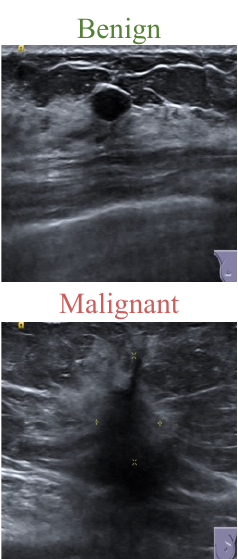}
    \subcaption{Target samples.}
    \label{fig:tsne_d}
\end{subfigure}
\vspace{-.5em}
\caption{%
\textbf{Failure Case Analysis.} We illustrate the t-SNE \cite{t-SNE} feature space of the black-box downstream model on the UDIAT$\rightarrow$UCLM task. 
The analysis is presented under three settings: (a) before style transfer, (b) with domain-level alignment only, and (c) with both domain- and class-level alignment.
We illustrate three failure cases: \textcolor{RoyalBlue}{easy}, \textcolor{orange}{medium}, and \textcolor{byzantine}{hard}, using the same samples across settings. The \textcolor{RoyalBlue}{easy} case is misclassified only before style transfer, the \textcolor{orange}{medium} case remains misclassified after domain-level alignment, and the \textcolor{byzantine}{hard} case persists under all settings. Meanwhile, by comparing the same sample across different settings,  we show the progressive influence of style transfer under different settings. \textit{Please zoom in for better visibility.}
} \label{fig:tsne_failure}
\vspace{-.5em}
\end{figure*}

\subsection{Failure Case Analysis} \label{subsec:failure_case}

We analyze failure cases within the feature space of the black-box downstream model using t-SNE \cite{t-SNE}, categorizing them into three cases—\textcolor{RoyalBlue}{easy}, \textcolor{orange}{medium}, and \textcolor{byzantine}{hard}—as shown in \cref{fig:tsne_failure}. For clarity, we further examine them under three settings:

\begin{enumerate}
    \item Setting 1 (\textbf{S1}): We denote the \textit{before style transfer} setting as no style transfer applied. As shown in \cref{fig:tsne_a}, the source and target domains remain misaligned.
    \item Setting 2 (\textbf{S2}): We introduce our pattern-matching module to alleviate the domain gap. We refer to this configuration as \textit{only domain level}, since the alignment focuses solely on transferring domain-specific appearance, as shown in \cref{fig:tsne_b}. 
    \item Setting 3 (\textbf{S3}): Finally, we simultaneously minimize both domain-level and class-level discrepancies through our proposed dual-level stylization module. This configuration is referred to as \textit{domain and class levels}, as shown in \cref{fig:tsne_c}. 
\end{enumerate}

In the \textcolor{RoyalBlue}{easy} case, the source sample (\textcolor{RoyalBlue}{blue}-bordered image) is initially misclassified in \textbf{S1}. In \textbf{S2}, the same sample successfully matches the appearance of the target data (see more \cref{fig:tsne_d} for the comparison), leading to a correct classification. Furthermore, this alignment continues improvements with \textbf{S3}, the sample moves further from the decision boundary, providing more robust predictions.

However, when we consider the \textcolor{orange}{medium} case (example by the \textcolor{orange}{orange}-bordered image), \textbf{S2} is insufficient to preserve class-discriminative properties (e.g., the tumor region highlighted in \textcolor{red}{red}-square \textcolor{red}{$\square$} of \cref{fig:tsne_b}), leading to ambiguous class confusion. In contrast, with \textbf{S3}, the benign-specific characteristics are preserved (see the \textcolor{red}{red}-square \textcolor{red}{$\square$} in \cref{fig:tsne_c}), which effectively drives the misclassified sample toward the correct class.

More critically, we observe the \textcolor{byzantine}{hard} case (shown by the \textcolor{byzantine}{purple}-bordered image), where the sample exhibits inherent differences in structure and tissue characteristics compared with the target data. As a result, even with \textbf{S3}, we still encounter a misclassification for this specific sample.

\section{Discussion} \label{sec:discussion}

\subsection{Can UI-Styler Achieve Scalability and Generalization?} \label{subsec:scal_gener}

\noindent \textbf{Scalability.}  
To demonstrate the scalability of UI-Styler in real-world deployments with multiple source domains, we explore two training strategies:  
\begin{enumerate}[leftmargin=2em, itemsep=0pt, topsep=0pt]
    \item \textit{Single}-source setting: the model is trained on one source domain (either BUSBRA or BUSI) and evaluated on the corresponding source$\rightarrow$UDIAT task.  
    \item \textit{Multi}-source setting: the model is trained jointly on (BUSBRA+BUSI)$\rightarrow$UDIAT and then evaluated on both source$\rightarrow$UDIAT tasks within a unified model, which alleviates the need for training $N \times (N-1)$ separate models as required by the \textit{single}-source setting, where $N$ denotes the number of devices.
\end{enumerate}
As shown in the \colorbox{teal!25}{seen} part of \cref{tab:multisource}, \textit{multi}-source training achieves performance comparable to \textit{single}-source training, with only a small gap (e.g., BUSBRA$\rightarrow$UDIAT AUC $71.52$ vs.\ $71.31$ and BUSI$\rightarrow$UDIAT Dice $80.49$ vs.\ $80.39$), while consistently outperforming the baseline without style transfer (w/o ST).

\noindent \textbf{Generalization.}  
We further evaluate the generalization ability of UI-Styler by selecting BUSBRA and BUSI as the seen source domains, UCLM as the unseen source domain, and keeping UDIAT as the fixed target.  
\begin{enumerate}[leftmargin=2em, itemsep=0pt, topsep=0pt]
    \item \textit{Single}-source setting: the model is trained on BUSBRA$\rightarrow$UDIAT and then evaluated on UCLM$\rightarrow$UDIAT.  
    \item \textit{Multi}-source setting: the model is trained jointly on (BUSBRA+BUSI)$\rightarrow$UDIAT and evaluated on UCLM$\rightarrow$UDIAT.  
\end{enumerate}
As shown in the \colorbox{red!25}{unseen} part of \cref{tab:multisource}, the single-source model already achieves solid performance, while the multi-source setting further improves results across multiple metrics, with Acc increasing from $65.00$ to $67.50$ and AUC from $70.32$ to $72.62$. These findings provide strong evidence of UI-Styler’s effectiveness in adapting to new, unseen devices in practical scenarios.

\begin{table}[!t]
\centering
\resizebox{\columnwidth}{!}{%
\begin{tabular}{@{\hspace{.1em}}c@{\hspace{.2em}}c|c|cccccc}
\toprule
\multicolumn{2}{c|}{Tasks} & Settings & KID\textcolor{red}{$\downarrow$} & Acc\textcolor{red}{$\uparrow$} & AUC\textcolor{red}{$\uparrow$} & Dice\textcolor{red}{$\uparrow$} & IoU\textcolor{red}{$\uparrow$} \\

\midrule

\multirow{6}{*}{\colorbox{teal!25}{\rotatebox{90}{Seen}}} & \multirow{3}{*}{\shortstack{BUSBRA\\$\downarrow$\\UDIAT}} & w/o ST & 13.81 & 55.95 & 64.29 & 84.76 & 75.71 \\
& & Single & \textbf{9.14} & \textbf{72.47} & \textbf{71.52} & \textbf{86.04} & \textbf{77.52} \\
& & Multi & \underline{12.24} & \underline{68.74} & \underline{71.31} & \underline{85.83} & \underline{76.93} \\

\cmidrule(lr){2-8}

& \multirow{3}{*}{\shortstack{BUSI\\$\downarrow$\\UDIAT}} & w/o ST & 7.23 & 73.33 & 73.16 & 79.53 & 70.61 \\
& & Single & \textbf{3.61} & \underline{74.36} & \textbf{78.89} & \textbf{80.49} & \textbf{71.61} \\
& & Multi & \underline{4.00} & \textbf{75.38} & \underline{78.43} & \underline{80.39} & \underline{71.34} \\

\midrule
\midrule

\multirow{3}{*}{\colorbox{red!25}{\rotatebox{90}{Unseen}}} & \multirow{3}{*}{\shortstack{UCLM\\$\downarrow$\\UDIAT}} & w/o ST & 20.90 & 63.75 & 68.15 & 82.22 & 72.06 \\

& & Single & \underline{10.84} & \underline{65.00} & \underline{70.32} & \textbf{82.71} & \textbf{72.64} \\
& & Multi & \textbf{9.67} & \textbf{67.50} & \textbf{72.62} & \underline{82.66} & \underline{72.57} \\

\bottomrule
\end{tabular}
}
\vspace{-.5em}
\caption{%
\textbf{Can UI-Styler Achieve Scalability and Generalization?} 
We assess scalability and generalization with BUSBRA and BUSI as the \textcolor{teal}{seen} source domains, UCLM as the \textcolor{red}{unseen} source domain, and UDIAT as the fixed target. 
In the \colorbox{teal!25}{seen} setting, models are trained \emph{and evaluated} on the corresponding source$\rightarrow$UDIAT tasks (single: one source; multi: BUSBRA+BUSI). 
In the \colorbox{red!25}{unseen} setting, models are trained on BUSBRA$\rightarrow$UDIAT (single) or (BUSBRA+BUSI)$\rightarrow$UDIAT (multi) and evaluated on UCLM$\rightarrow$UDIAT. 
w/o ST denotes training without style transfer.
} \label{tab:multisource}
\end{table}

\subsection{How Noisy Pseudo Target Labels Affect Performance?} \label{subsec:noisy}

Since pseudo target labels are generated by a black-box downstream model, \textit{label noise is an inevitable factor in realistic deployments}. 
To investigate the robustness of UI-Styler against noisy labels, we conduct experiments on the BUSI$\rightarrow$BUSBRA task by progressively injecting noise from $0\%$ to $40\%$ into the target domain. Specifically, we randomly replaced the ground truths with incorrect classes.

As shown in \cref{tab:noisy}, we observe that introducing a mild noise level of $10\%$ keeps the results almost unchanged compared to the clean setting ($0\%$). Even higher noise levels ($20$–$30\%$) lead to only \textbf{marginal} degradation across most metrics (e.g., AUC drops only slightly to $87.87$ and $87.61$), while all metrics continue to surpass the baseline without style transfer (w/o ST). These findings indicate that UI-Styler can tolerate moderate noise levels without noticeable performance loss. Only at $40\%$ noise, we observe a more visible decline, with AUC reduced to $86.77$ and Dice to $82.39$, yet UI-Styler still surpasses the w/o ST baseline on $3/5$ metrics (KID, Acc, and IoU).

These findings suggest that although UI-Styler does not incorporate any explicit noise-mitigation module, its design exhibits a certain degree of robustness to label noise. We acknowledge that heavy noise can accumulate errors through the proposed losses ($\mathcal{L}_{\text{dir}}$ and $\mathcal{L}_{\text{sup}}$), which may limit reliability in extreme cases. Nonetheless, the \textbf{stability under low-to-moderate noise} demonstrates that UI-Styler can operate effectively in realistic settings where the black-box downstream model achieves at least $70\%$ accuracy.

\textbf{Obviously}, black-box downstream models \textit{must} achieve accuracy well above $70$\% to be meaningful in medical applications. Models falling below this accuracy level are essentially random in outcome and often biased toward a single class. Consequently, their predictions are unsafe for diagnosis and provide clinicians with no reliable basis for decision-making.

\begin{table}[!t]
\centering
\resizebox{\columnwidth}{!}{%
\begin{tabular}{c|c||cccccc}
\toprule
Task & Noisy Levels & KID\textcolor{red}{$\downarrow$} & Acc\textcolor{red}{$\uparrow$} & AUC\textcolor{red}{$\uparrow$} & Dice\textcolor{red}{$\uparrow$} & IoU\textcolor{red}{$\uparrow$} \\

\midrule

\multirow{6.2}{*}{\shortstack{BUSI\\$\downarrow$\\BUSBRA}} & w/o ST & 19.73 & 82.56 & 87.30 & 82.41 & 73.37 \\

\cmidrule{2-7}

& 0\% & 11.25 & \textbf{85.13} & \textbf{88.14} & \textbf{83.15} & \textbf{74.05} \\

& 10\% & 11.20 & \textbf{85.13} & \underline{87.93} & \underline{82.92} & \underline{73.97} \\

& 20\% & \textbf{11.14} & \underline{84.10} & 87.87 & 82.70 & 73.70 \\

& 30\% & \underline{11.19} & 83.59 & 87.61 & 82.68 & 73.67 \\

& 40\% & 11.26 & 83.08 & 86.77 & 82.39 & 73.45 \\

\bottomrule
\end{tabular}%
}
\vspace{-.5em}
\caption{%
\textbf{How Noisy Pseudo Target Labels Affect Performance?} 
We report results on the BUSI$\rightarrow$BUSBRA task under different noise levels ($0\%$, $10\%$, $20\%$, $30\%$, and $40\%$), where noise is introduced by randomly replacing ground truths with incorrect class assignments. 
Even with $40\%$ noisy labels, UI-Styler still surpasses the baseline without style transfer (w/o ST) on $3/5$ metrics (KID, Acc, and IoU).
} \label{tab:noisy}
\end{table}



\section{Cross-device Visual Results} \label{sec:visual}

To further assess the effectiveness of the proposed UI-Styler, we present visual results for all $12$ source-to-target transfer tasks, alongside representative examples that highlight the unique appearance characteristics of each ultrasound dataset, as shown in Fig.~\ref{tab:visual_all}.
Each subfigure corresponds to a specific domain adaptation scenario, where the top row shows target domain samples, the middle row displays source domain inputs, and the bottom row presents the stylized outputs produced by UI-Styler.

Visually, UI-Styler consistently adapts the source image style to match the target domain while preserving tumor structure and lesion boundaries. The translated images demonstrate improved textural consistency and contrast characteristics aligned with the target domain, including probe artifacts, intensity ranges, and noise profiles. Notably, the stylized outputs retain key diagnostic features essential for downstream classification and segmentation tasks.

Beyond enhancing model performance, this visual consistency also supports clinical interpretation. By translating unfamiliar input styles into the target domain's appearance, UI-Styler facilitates diagnostic reasoning for physicians, especially when deploying models trained on known devices to new acquisition environments. This alignment reduces adaptation burden and promotes safe model deployment in device-diverse clinical settings.

\pagebreak

\newpage
\begin{figure*}[ht]
\centering

\includegraphics[width=\textwidth]{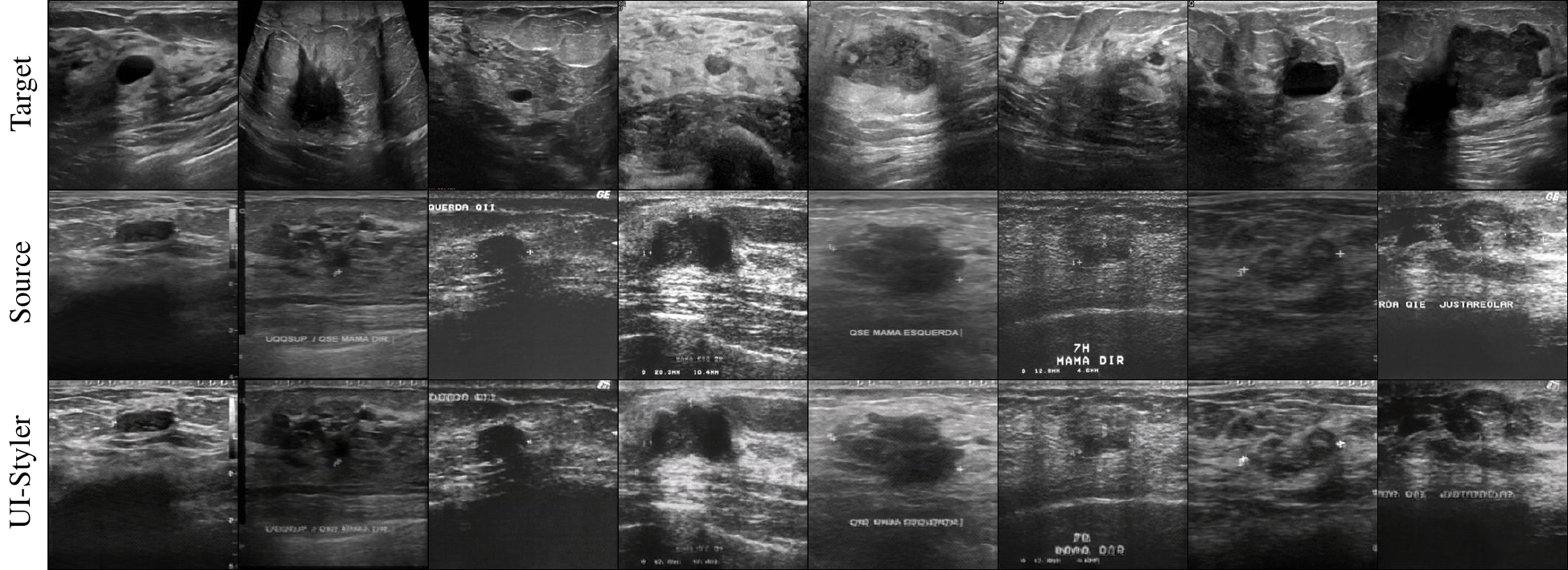}
\small (a) BUSBRA$\rightarrow$BUSI. \vspace{1em}

\includegraphics[width=\textwidth]{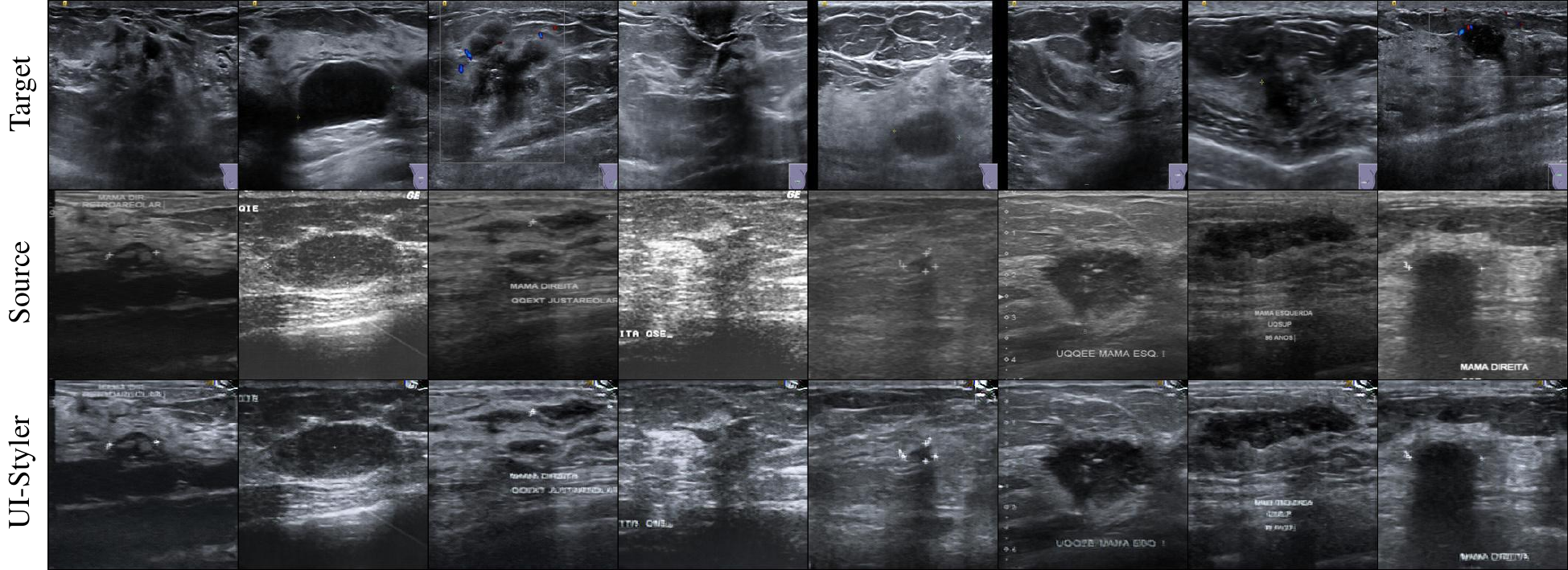}
\small (b) BUSBRA$\rightarrow$UCLM. \vspace{1em}

\includegraphics[width=\textwidth]{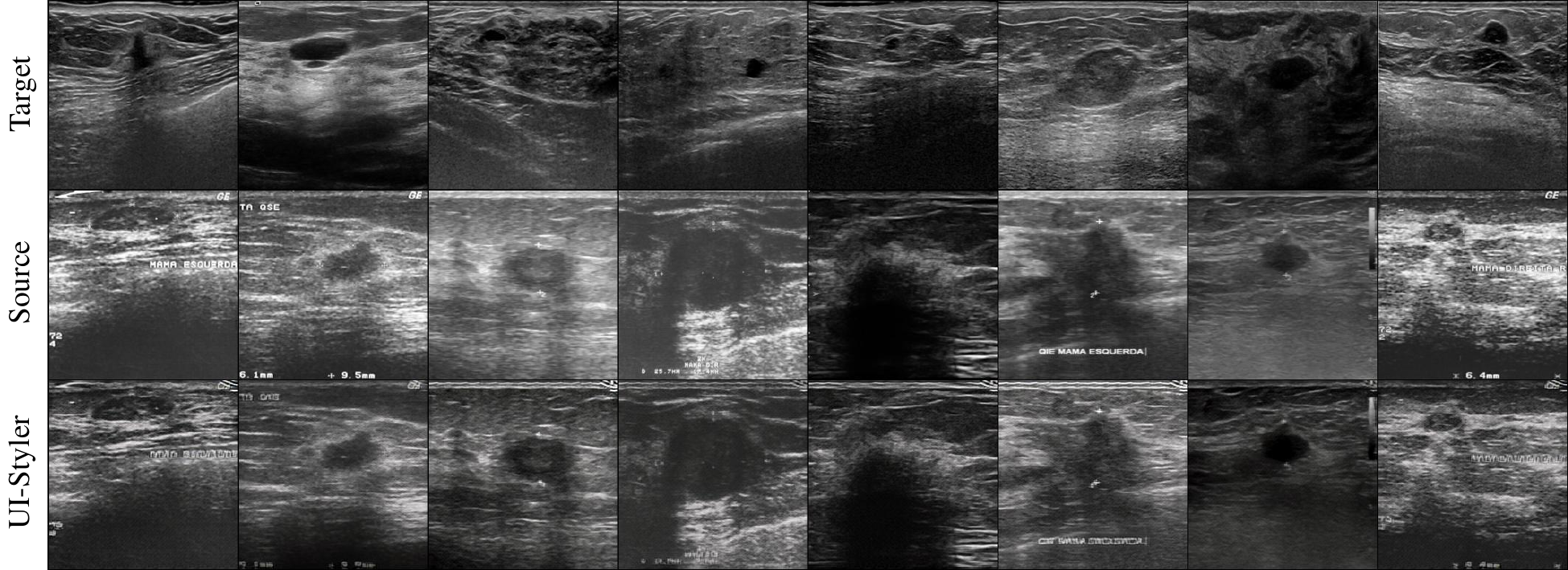}
\small (c) BUSBRA$\rightarrow$UDIAT. \vspace{1em}

\end{figure*}

\newpage
\begin{figure*}[ht]
\centering

\includegraphics[width=\textwidth]{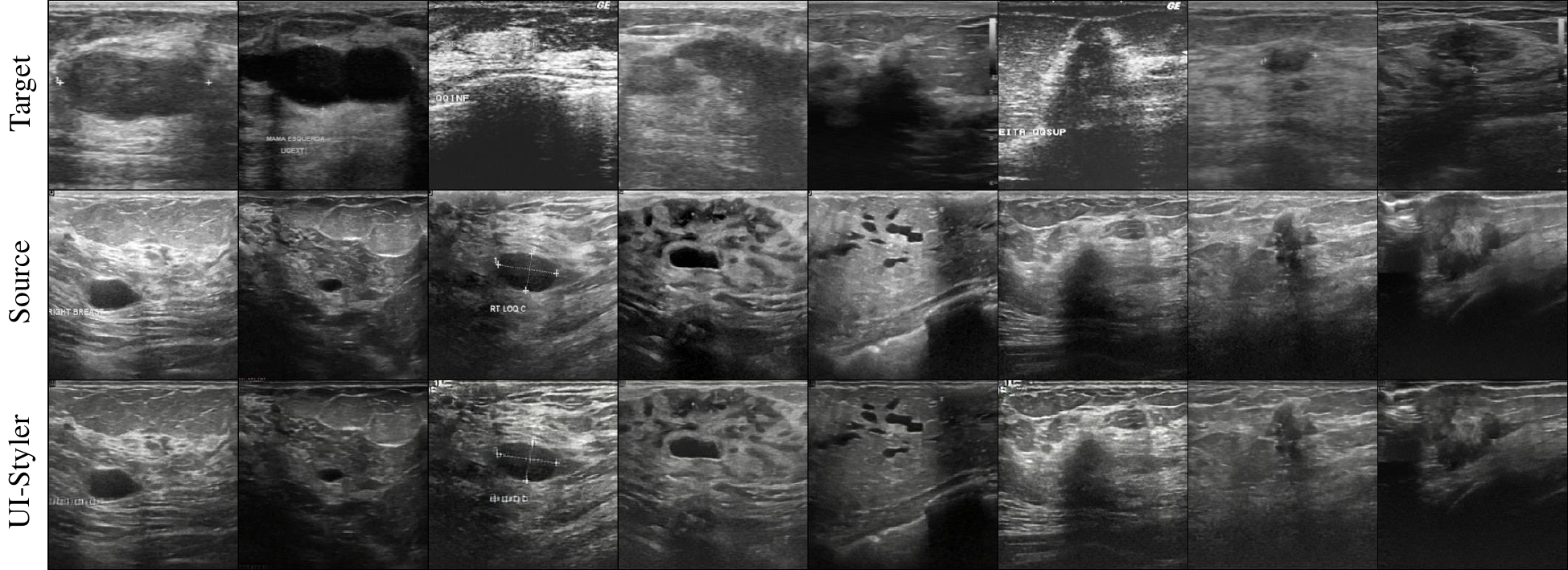}
\small (d) BUSI$\rightarrow$BUSBRA. \vspace{1em}

\includegraphics[width=\textwidth]{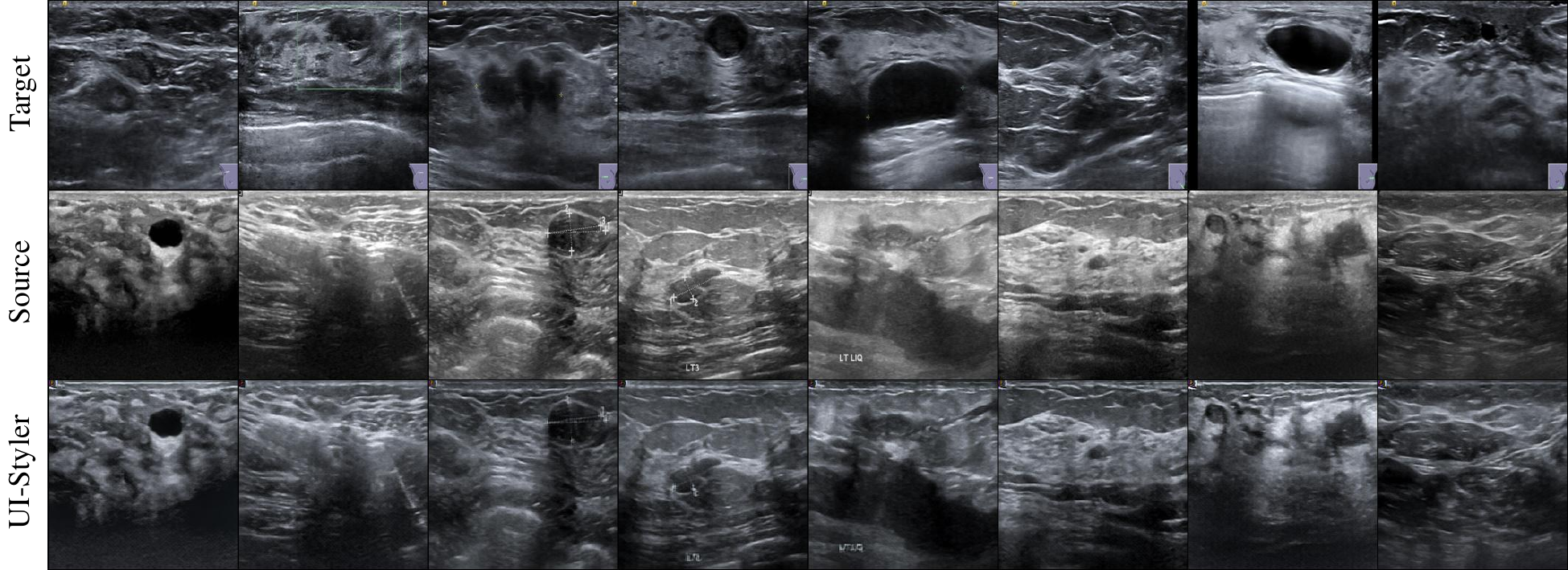}
\small (e) BUSI$\rightarrow$UCLM. \vspace{1em}

\includegraphics[width=\textwidth]{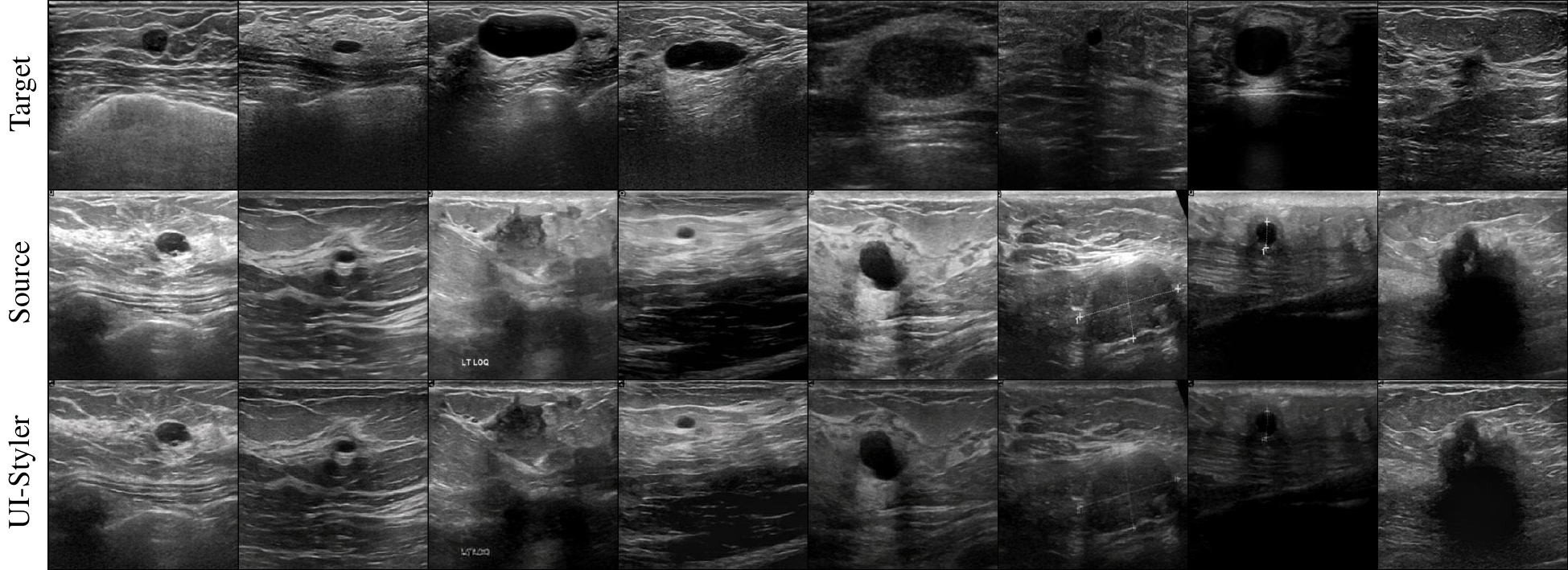}
\small (f) BUSI$\rightarrow$UDIAT. \vspace{1em}

\end{figure*}

\newpage
\begin{figure*}[ht]
\centering

\includegraphics[width=\textwidth]{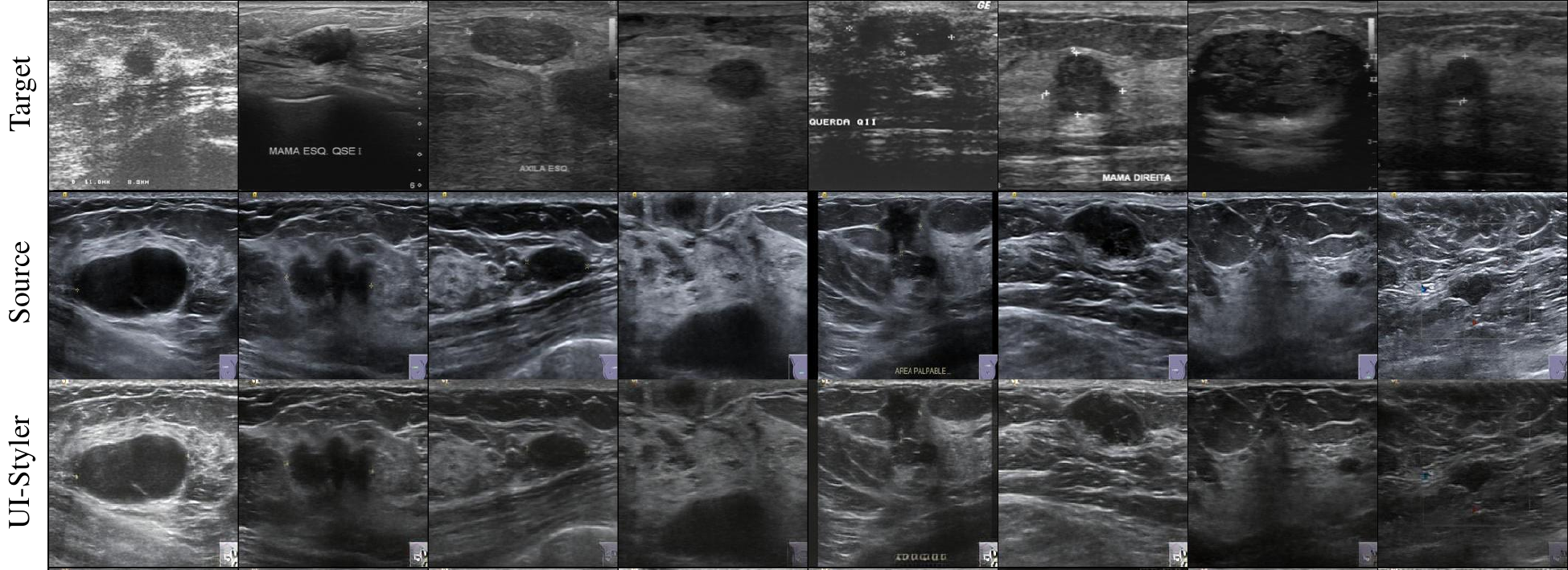}
\small (g) UCLM$\rightarrow$BUSBRA. \vspace{1em}

\includegraphics[width=\textwidth]{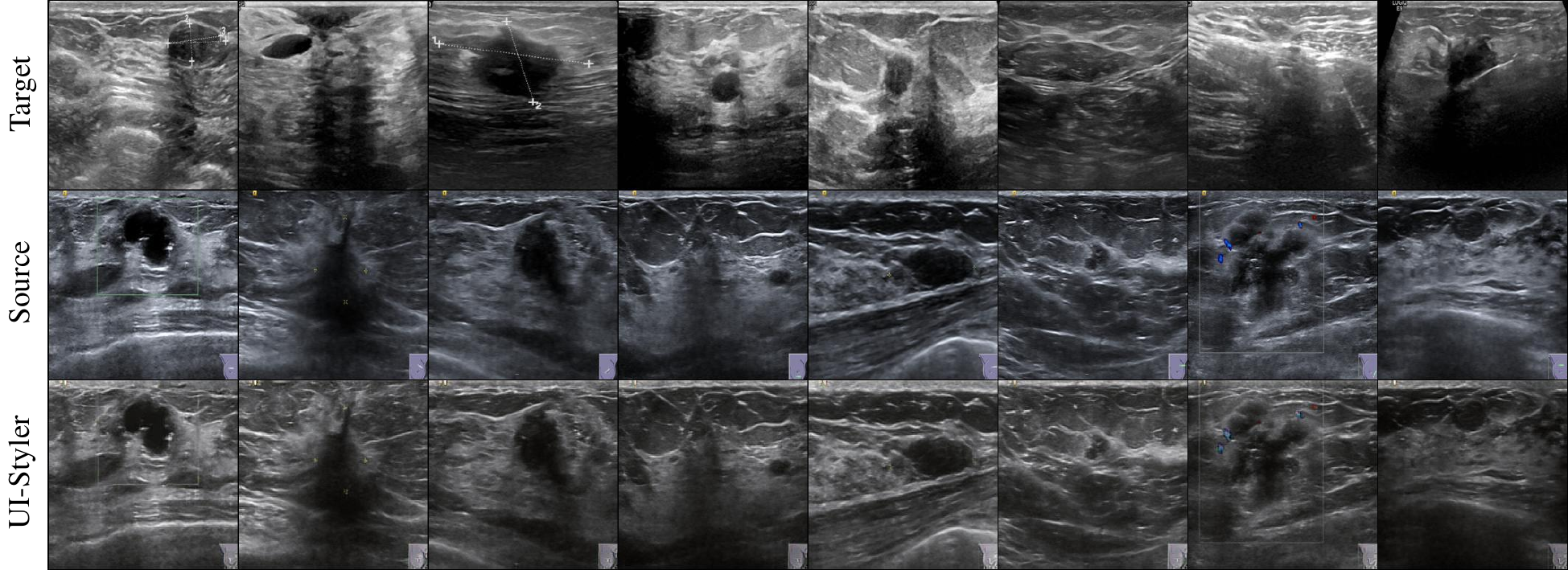}
\small (h) UCLM$\rightarrow$BUSI. \vspace{1em}

\includegraphics[width=\textwidth]{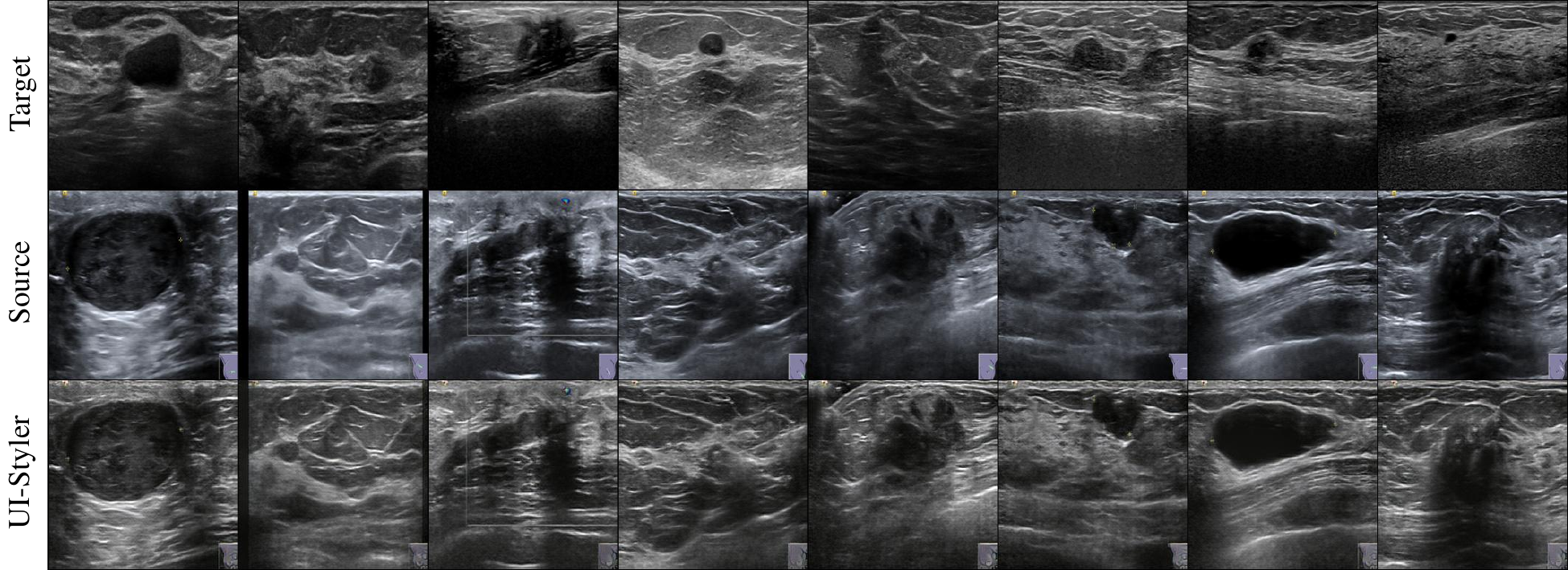}
\small (i) UCLM$\rightarrow$UDIAT. \vspace{1em}

\end{figure*}

\setcounter{figure}{3}
\setcounter{table}{3}

\newpage
\begin{figure*}[t]
\centering

\includegraphics[width=\textwidth]{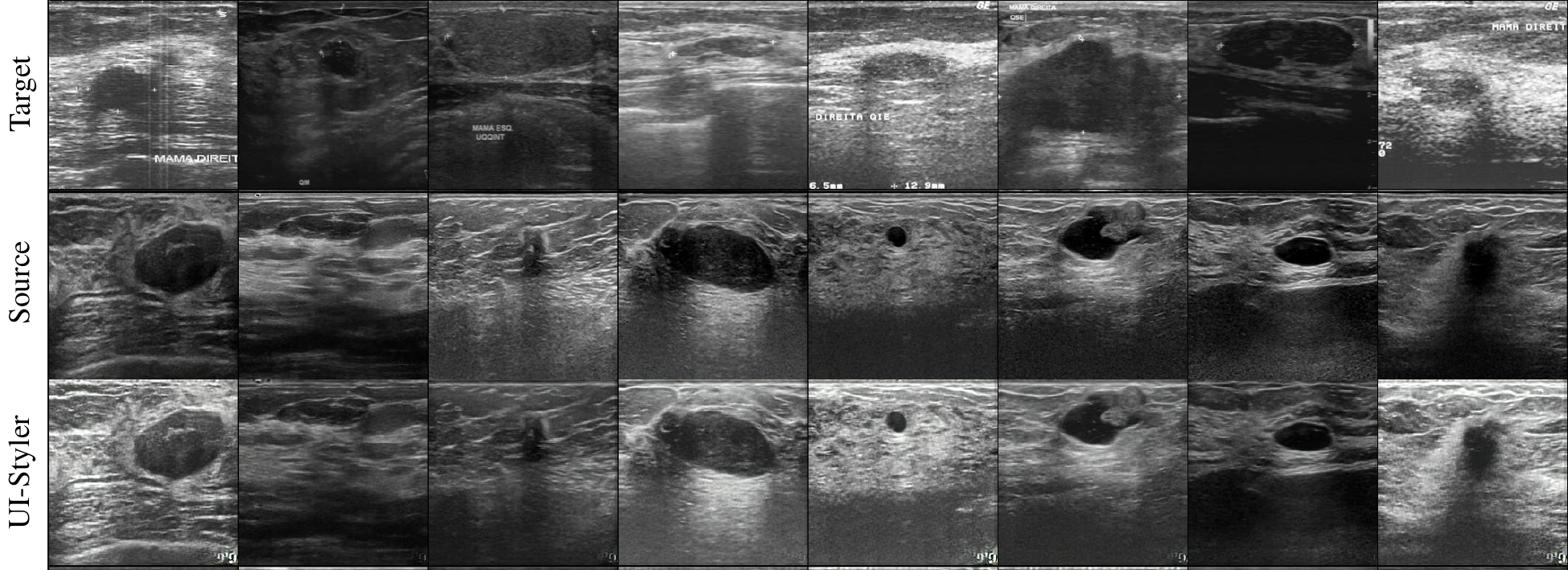}
\small (j) UDIAT$\rightarrow$BUSBRA.

\includegraphics[width=\textwidth]{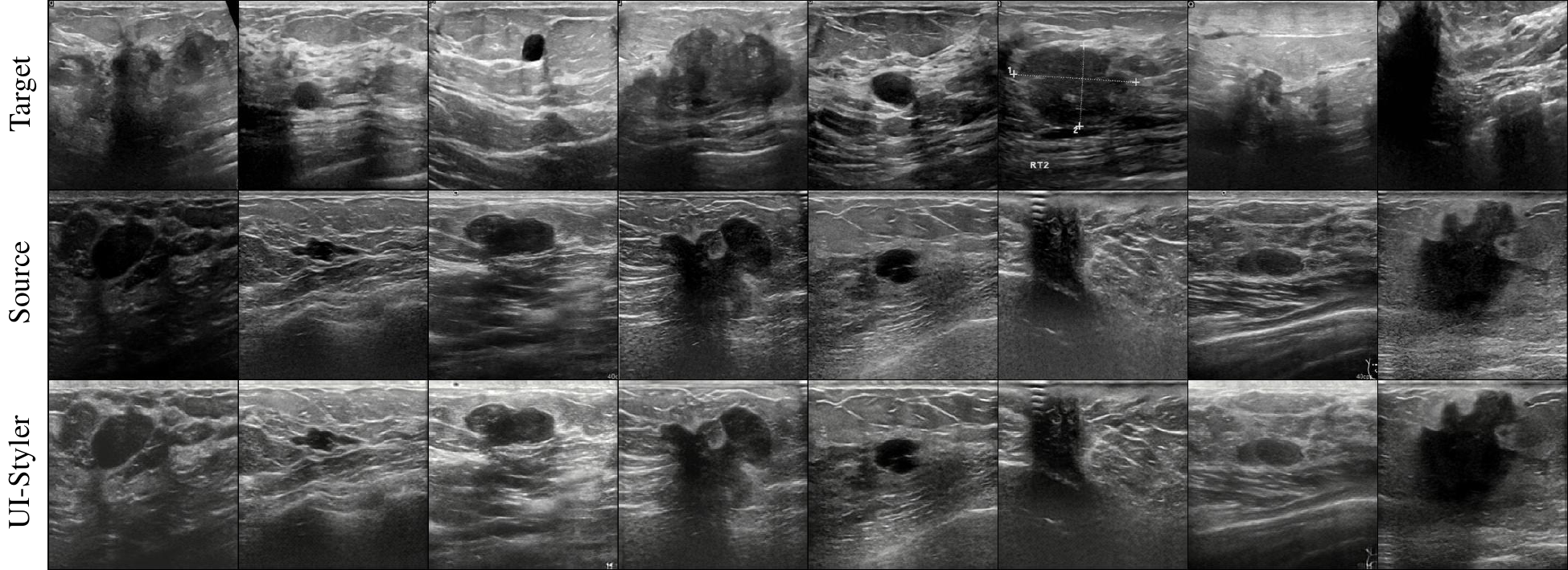}
\small (k) UDIAT$\rightarrow$BUSI.

\includegraphics[width=\textwidth]{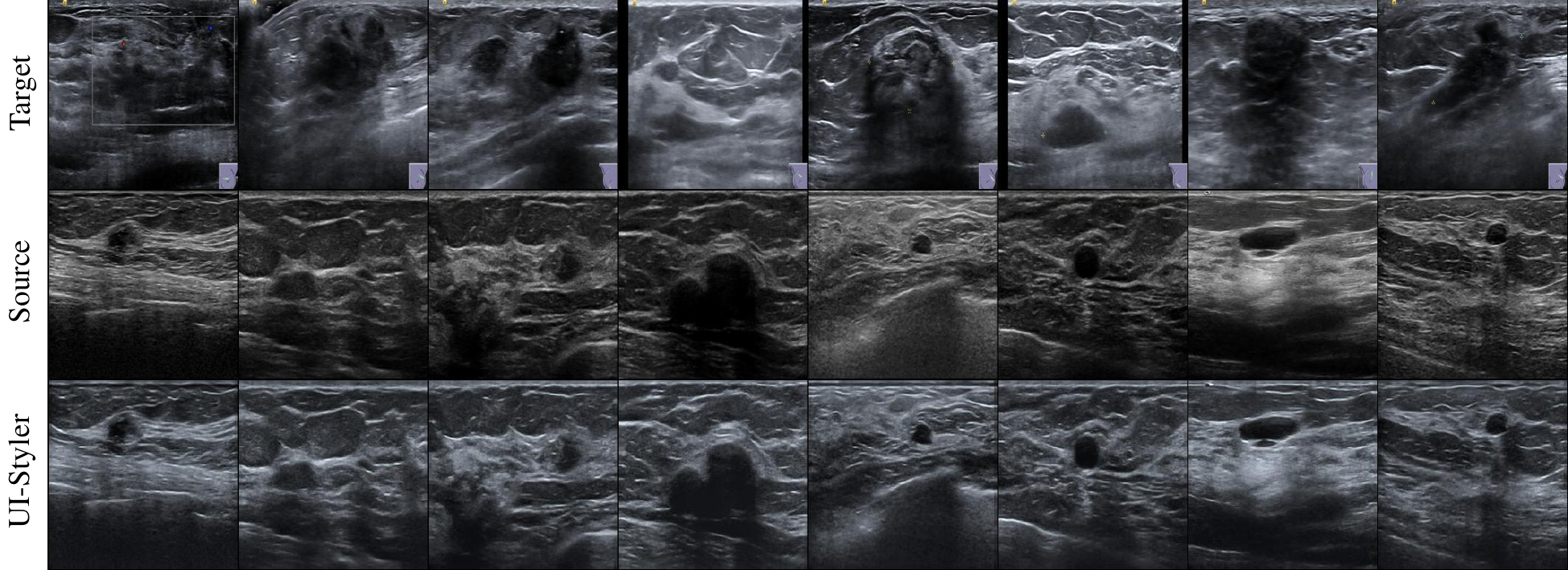}
\small (l) UDIAT$\rightarrow$UCLM.

\caption{
\textbf{Cross-device Visual Results.}
We present qualitative results of UI-Styler across all $12$ cross-device ultrasound translation tasks.
Each group shows representative examples from the target domain (top), source domain (middle), and the stylized results by UI-Styler (bottom).
}
\label{tab:visual_all}

\end{figure*}

\pagebreak
{
    \small
    \bibliographystyle{ieeenat_fullname}
    \bibliography{main}
}